\documentclass{article}

    \usepackage[final]{neurips_2022}

\usepackage[utf8]{inputenc} 
\usepackage[T1]{fontenc}    
\usepackage{graphicx}
\usepackage{caption}
\usepackage{subcaption}
\usepackage{hyperref}       
\usepackage{url}            
\usepackage{booktabs}       
\usepackage{amsmath,amsfonts,amssymb}      
\usepackage{nicefrac}       
\usepackage{microtype}      
\usepackage{xcolor}         
\usepackage{placeins}
\usepackage{natbib}
\usepackage{enumitem}
\title{An Empirical Analysis of the Advantages of \\
Finite- v.s.~Infinite-Width Bayesian Neural Networks}

\author{%
  Jiayu Yao \quad Yaniv Yacoby \quad Beau Coker \quad Weiwei Pan \quad Finale Doshi-Velez \\
    Harvard University \\
  \texttt{\{jiy328, yanivyacoby, beaucoker, weiweipan, finaledoshivelez\}@g.harvard.edu} \\
}

\begin{document}

\maketitle
\begin{abstract}
Comparing Bayesian neural networks (BNNs) with different widths is challenging because, as the width increases, multiple model properties change simultaneously, and, inference in the finite-width case is intractable. In this work, we  empirically compare finite- and infinite-width BNNs, and provide quantitative and qualitative explanations for their performance difference.
We find that when model is mis-specified, increasing width can hurt BNN performance. 
In these cases, we provide evidence that finite-width BNNs generalize better partially due to the properties of their frequency spectrum that allows them to 
adapt under model mismatch.
\end{abstract}

\section{Introduction}
Recent works in Bayesian deep learning note a counter-intuitive phenomenon---that larger architectures (specifically larger width) can hurt model performance. For example, ~\citeauthor{lee2017deep} observe that on image classification tasks, as the width of the Bayesian CNN increases, classification accuracy decreases; ~\citeauthor{pleiss2021limitations} make similar observations on regression tasks. 
Despite better performance of smaller NNs, posterior inference in finite-width BNNs is challenging~\citep{coker2022wide,foong2020expressiveness},
whereas, in the infinite-width limit, BNNs converge to a well-behaved model, the Neural Network Gaussian Processes (NNGPs) (e.g., \citep{lee2017deep,matthews2018gaussian,neal2012bayesian}). 
Furthermore, infinite-width BNNs easily admit theoretical analysis while finite-width BNNs are difficult to analyze due to the non-Gaussian processes~\citep{noci2021precise} they define in function space. 
Thus, to choose between finite- versus infinite-width BNNs, we need to know: when and why does solving the harder inference problem present a better trade-off for improved performance? 
In this paper, we provide empirical explanations for the counter-intuitive phenomenon that larger width hurts BNN performance, by comparing the inductive biases and frequency spectra of finite- and infinite-width BNNs.

\textbf{Contributions} 
We study one-hidden-layer BNNs with erf and ReLU activations, whose infinite-width counterparts are referred to as limiting NNGPs in the following. Our contributions are\footnote{Code Base: \href{https://github.com/dtak/finite_vs_infinite_bnn_public}{https://github.com/dtak/finite\_vs\_infinite\_bnn\_public}}: 
\begin{enumerate}[leftmargin=0.5cm]
\item We show that when there is model mismatch between the limiting NNGP and the data generating model, finite-width BNNs can generalize better than infinite-width BNNs;

\item To study the above phenomenon,
we quantitatively show that finite-width BNNs have different inductive biases from NNGPs (finite-width BNNs generate more diverse datasets than NNGPs);

\item We also qualitatively compare the inductive biases of finite- and infinite-width BNNs by comparing their frequency spectra from spectral analysis. 
We demonstrate that, compared to limiting NNGPs, the prior predictive distributions of finite-width BNNs define heavier-tailed coefficient distributions over the frequency spectrum (they place more mass on large spectral coefficients). Additionally, after inference, the posterior predictive distributions of finite-width BNNs define frequency spectra that are more similar to those defined by the data generating model;

\item We hypothesize that because finite-width BNNs define frequency spectra with heavy-tailed spectral coefficient distributions, they are more flexible in modeling the data using different frequencies. 
To test this, we design an inference procedure to filter out high-frequency components from the functional prior of BNNs.
We see that the generalization performance of BNNs decreases as more high-frequency components are removed from the function, supporting our hypothesis.
Additionally, we reconstruct datasets by removing the high-frequency components. We observe that finite-width BNNs outperform the limiting NNGPs with their spectral coefficients of high-frequency components more concentrated at $0$, which resembles the true data generating process.
\end{enumerate}

In this work, we consider both erf and ReLU nonlinearity, since in these cases we obtain closed-form kernels for the limiting NNGPs, which allows for exact inference and theoretical analysis. 
Additionally, both activation functions are well studied in the literature (e.g., \citep{williams2006gaussian,cho2011analysis}) and are suitable for modeling various types of functions~\citep{lee2020neural}. 

%
\textbf{Related Work} 
A large body of work studies when and why finite-width (non-Bayesian) NNs succeed. For example,~\citeauthor{ghorbani2020neural} and~\citeauthor{refinetti2021classifying} empirically find that finite-width NNs outperform kernel methods when the data contains low-dimensional structures. \cite{aitchison2020bigger} shows that finite deep linear networks yield stochastic kernels, which are more flexible in learning representations from the data.
However, it is not known whether these insights apply to BNNs.

In the context of BNNs, 
\citeauthor{pleiss2021limitations} study the performance degeneration of infinite-width BNNs by studying deep GPs, which the authors consider to be a more generalized BNN model.
The authors prove that as the width of a deep GP increases, the posterior collapses from a mixture function of data-dependent bases to that of data-independent bases. 
Thus, infinitely wide deep GPs are less capable of adapting to the data. Similar insights have been made in~\cite{aitchison2021deep}.
We study BNNs from a spectral perspective and find that finite-width BNNs are more flexible in that they can use functions of different frequencies.

Analogously to standard NNs~\citep{rahaman2019spectral}, existing works observe that, unlike their infinite counterparts, finite-width BNNs tend to be heavy-tailed in function space (e.g., \citep{noci2021precise,pleiss2021limitations}).  
We examine how and why this heavy-tailedness affects the prediction tasks under model mismatch. 

Finally, a recent line of work studies the spectral properties and the reproducing kernel Hilbert space of infinite-width NNs (e.g., \citep{chen2020deep,bietti2020deep}).
Another body of work provides theoretical and empirical analyses of how the power spectrum of standard, wide NNs changes during training (e.g., \citep{rahaman2019spectral, cao2019towards, yang2019fine}).  However, both bodies of work do not consider the finite-width, Bayesian settings, nor the relationship between the power spectrum and generalization performance.

\textbf{Background}
\label{sec:background}
We consider a regression task with dataset $\mathcal{D} = \{\mathbf{x}_i,y_i\}_{i=1}^n$, $\mathbf{x}_i\in \mathbb{R}^{d_{\text{in}}}$, $ y_i \in \mathbb{R}$. For an NN with one hidden-layer of width $H$, the  output is given by $f_\text{NN}(\mathbf{x}_i) = \frac{1}{\sqrt{H}}\mathbf{w}_1 \phi(\mathbf{w}_0 \mathbf{x}_i+\mathbf{b}_0)+ b_1$,
with $\mathbf{w}_0\in\mathbb{R}^{H\times d_{\text{in}}},\mathbf{b}_0\in\mathbb{R}^{H\times 1},\ \mathbf{w}_1\in\mathbb{R}^{1\times H}, b_1\in\mathbb{R}$, and $\phi$ is a nonlinear activation function.

We denote neural network weights as $\mathbf{W}$ and biases as $\mathbf{b}$. For a Bayesian Neural Network, we specify prior distributions $p(\mathbf{W})\sim\mathcal{N}(0,\sigma_\mathbf{W}^2 \mathbf{I} ),\ p(\mathbf{b})\sim\mathcal{N}(0,\sigma_\mathbf{b}^2 \mathbf{I} )$.
The posterior distribution of the weights is given by:
$p(\mathbf{W},\mathbf{b}|\mathcal{D})\propto p(\mathcal{D}|\mathbf{W},\mathbf{b})p(\mathbf{W})p(\mathbf{b})$,
where $p(\mathcal{D}|\mathbf{W},\mathbf{b}) = \prod_i p(y_i|\mathbf{x}_i,\mathbf{W},\mathbf{b}),\ p(y_i|\mathbf{x}_i,\mathbf{W},\mathbf{b}) = \mathcal{N}(y_i|f_\text{NN}(\mathbf{x}_i),\sigma^2_\epsilon)$ and $\sigma^2_\epsilon$ is the variance of the output noise. 

Given a set of new inputs $\mathcal{D}^* = \{\mathbf{x}^*_i\}_{i=1}^m$, the posterior predictive distribution is given by
\begin{equation*}
    p(f^*|\mathbf{x}^*,\mathcal{D}) = \int_{\mathbf{W},\mathbf{b}} p(f^*|\mathbf{x}^*,\mathbf{W},\mathbf{b},\mathcal{D}) p(\mathbf{W},\mathbf{b}|\mathcal{D}) d\mathbf{W}d\mathbf{b},\ \text{   where   }\ f^*=f_\text{NN}(\mathbf{x}^*).
\end{equation*}
We use NUTS~\citep{hoffman2014nuts} for sampling (Appendix~\ref{apdx:sec_inference}). With erf or ReLU activations, a BNN will converge to a GP with Arcsin or Arccos kernel, respectively (Appendix~\ref{apdx:sec_nngp_kernel})~\citep{williams2006gaussian,cho2011analysis}.
In practice, Arccos kernel has eigenvalues that decay slower than those of Arcsin kernel and is thus more suitable for learning rough functions~\citep{lee2020neural}.

\section{Generalization Performance of Finite- and Infinite-Width BNNs}

We compare the regression performance of finite- and infinite-width BNNs under model mismatch---we train BNNs whose limiting NNGPs are different from the data generating model. Thus, the performance of the limiting NNGP is not optimal, leaving room for improvement. 
We want to see if, under this model mismatch, finite-width BNNs perform differently from infinite-width BNNs.

\textbf{Datasets} We generate data using GPs with three  kernels: RBF, Arcsin and Arccos (GP-Arccos generates rougher functions that require a larger model capacity to learn). For GP-RBF, the lengthscale $l$ is chosen from $\{0.5, 1.0, 1.5 ,2.0\}$. For GP-Arcsin/Arccos, the hyperparameters $\sigma^2_{\mathbf{W}}, \sigma^2_{\mathbf{b}}$ in Equation~\ref{apdx:eqn_arcsin},~\ref{apdx:eqn_arccos}, are chosen from $\{0.5, 1.0, 1.5 ,2.0\}$.
For each kernel setting, we draw $200$ functions, each as a univariate dataset  (details in Appendix~\ref{apdx:sec_datasets}). We then fit each dataset using BNNs with erf and ReLU nonlinearity whose prior variances $\sigma^2_{\mathbf{W}}, \sigma^2_{\mathbf{b}}$ are chosen from $\{0.5, 1.0, 1.5 ,2.0\}$ and hidden widths $H$ vary from $2$ to $200$. For data drawn from a GP-RBF, there is always model-mismatch between the limiting NNGP and the data generating model.
For data drawn from a GP-Arcsin/Arccos, when the hyperparameters are equivalent to the prior variances of BNNs with erf/ ReLU nonlinearity, there is no model mismatch. 

\textbf{Evaluation Metrics}
To quantify the performance of finite-width BNNs, we calculate the expected difference of test negative loglikelihood (NLL) and test mean squared error (MSE). Specifically, for each kernel setting with datasets $\{\mathcal{D}^{(s)}\}_{s=1}^{S}$, we calculate
\begin{equation}\label{eqn:delta_ll}
    \Delta \text{NLL}_{\text{BNN}_H} = \frac{1}{S}\sum_{s=1}^{S} [\text{NLL}_{\text{BNN}_H}(\mathcal{D}^{(s)}) - \text{NLL}_\text{NNGP}(\mathcal{D}^{(s)})],
\end{equation} 
\begin{equation}\label{eqn:delta_mse}
    \Delta \text{MSE}_{\text{BNN}_H} = \frac{1}{S}\sum_{s=1}^{S} [\text{MSE}_{\text{BNN}_H}(\mathcal{D}^{(s)}) - \text{MSE}_\text{NNGP}(\mathcal{D}^{(s)})].
\end{equation}
When there is no model mismatch, both terms will converge to $0$ as hidden width $H$ goes to infinity.

\begin{figure*}[htbp]
     \centering
     \begin{subfigure}[b]{0.49\textwidth}
         \centering
         \includegraphics[width=\textwidth]{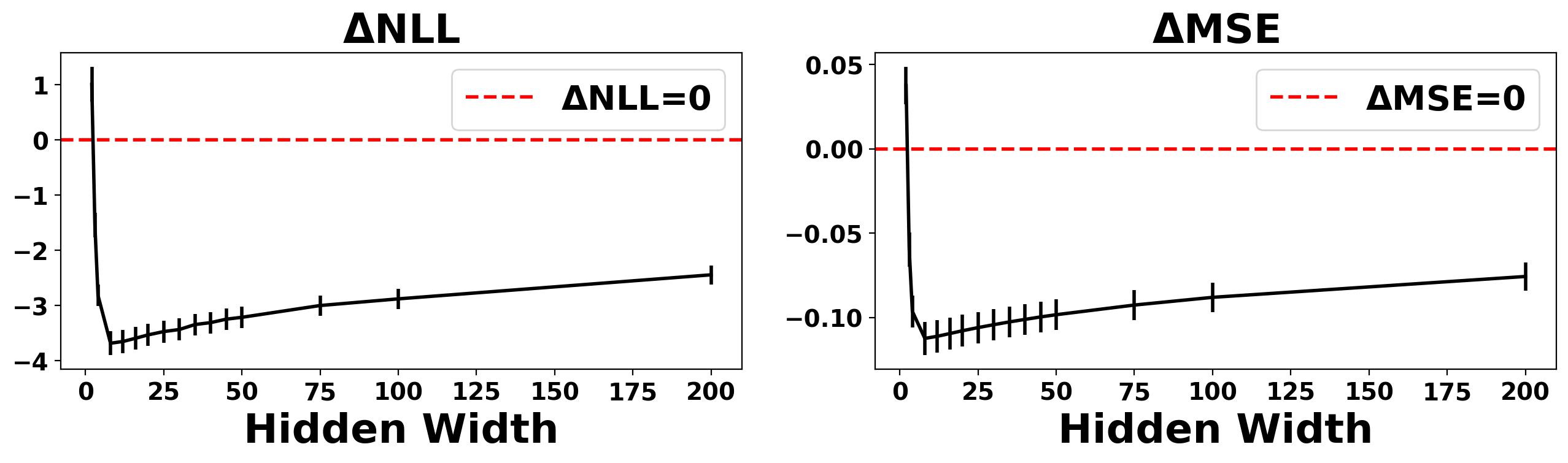}
         \caption{BNN with ReLU, $\sigma_\mathbf{W}^2=\sigma_\mathbf{b}^2=2.0$}
         \label{data_fit_a}
     \end{subfigure}
     \hfill
    \begin{subfigure}[b]{0.49\textwidth}
         \centering
         \includegraphics[width=\textwidth]{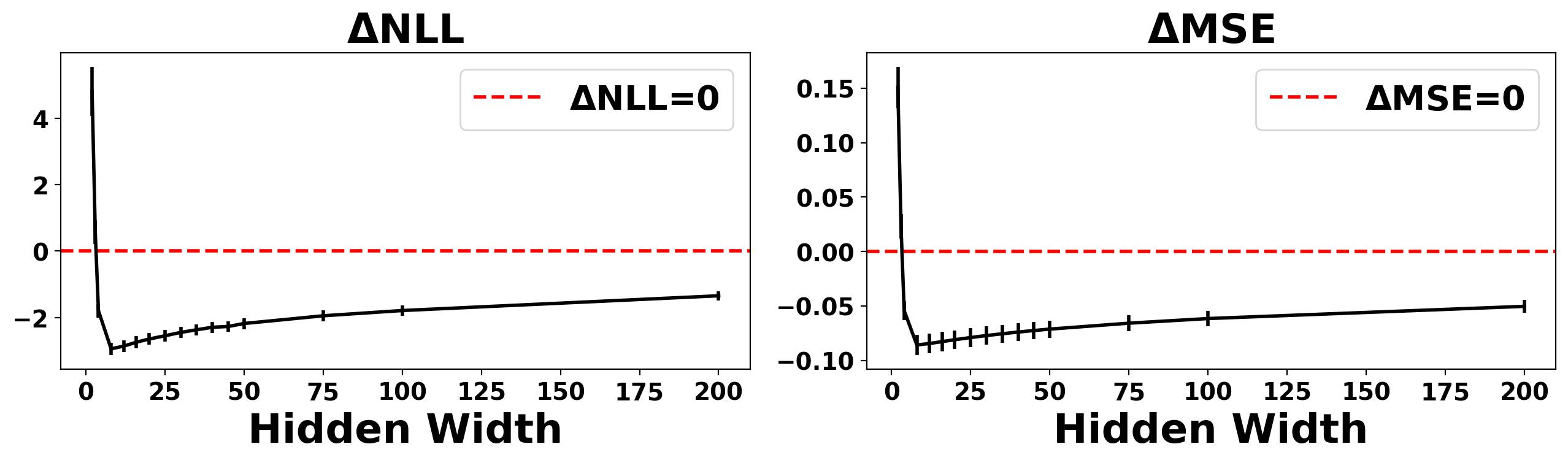}
         \caption{BNN with erf, $\sigma_\mathbf{W}^2=\sigma_\mathbf{b}^2=2.0$}
         \label{data_fit_b}
     \end{subfigure}
     \caption{The test performance of finite-width BNNs:  The data is generated from GP-RBF($l=0.5$). We fit each dataset with BNNs with erf and ReLU nonlinearity. 
     $x$-axis denotes the hidden-layer width of the BNN; $y$-axis denotes 
     $\Delta \text{NLL}$, $\Delta \text{MSE}$ (Equation~\ref{eqn:delta_ll},\ref{eqn:delta_mse}).
     Being above/below the red dashed line means that finite-width BNNs perform worse/better than the limiting NNGPs.
     We see that finite-width BNNs can outperform the limiting NNGPs as the black curves are mostly below the red dashed line.}
     \label{fig:data_fit}
\end{figure*}

\textbf{Results}
\textbf{When there is a model mismatch between the data generating model and the limiting NNGP, finite-width BNNs can outperform NNGPs.} Previous work observed similar phenomena on real datasets (e.g.,\cite{pleiss2021limitations,aitchison2021deep}), which inevitably have model mismatch. In Figure~\ref{data_fit_a},~\ref{data_fit_b}, we see that finite-width BNNs have better generalization performance (the black curves are below the red dashed lines) and converge to the limiting NNGP when the hidden width increases (the black curves approach the red dashed lines). 
We see similar trends on large real datasets (Figure~\ref{fig:uci}). 
Additionally, a larger model mismatch increases the difference in the generalization performance (Figure~\ref{fig:appdx_arccos_0.5},~\ref{fig:appdx_arcsin_0.5}). 

\section{Inductive Biases of Finite- and Infinite-Width BNNs}
Why do finite-width BNNs outperform infinite-width BNNs even though they have a smaller model capacity? To answer this question, we first empirically show that finite-width BNNs have different inductive biases from infinite-width BNNs (finite- and infinite-width BNNs are two distinct model classes); thus, one may generalize better than the other given different situations.

\subsection{Quantitatively Comparing Inductive Biases  Using the Data Likelihood}
There are different ways to quantify inductive biases, one of which is to evaluate  how much the model assumptions  overlap with the true data generating process.
In the Bayesian literature, log marginal likelihood (LML) is often used to quantify this overlap. 
The LML of generating dataset $\mathcal{D}$ under model $\mathcal{M}$ can be computed by marginalizing out the prior over the function values $f$,
\begin{equation}
\label{eqn:lml}
    \text{LML} = \log p(\mathcal{D}|\mathcal{M}) = \log \int p(\mathcal{D}|\mathcal{M},f)p(f|\mathcal{M})df.
\end{equation}
In our setup, the dataset $\mathcal{D}$ is generated from a GP and $f$ is modeled by either a BNN or an NNGP. 

Equation~\ref{eqn:lml} quantifies the inductive bias (overlap between the model assumptions and the data generating process) by answering the question ``\textit{how likely is the dataset to be generated by the model}?'' The literature argues that a model has a good inductive bias when it can achieve high LML for various datasets (a model is too complex when achieving modest LML for many datasets and too simple with high LML for limited datasets (e.g., \citep{mackay1992bayesian})).  Additionally, higher LML does not necessarily imply better generalization performance (e.g., \citep{lotfi2022bayesian}).

Despite many good properties of LML, Equation~\ref{eqn:lml} is hard to compute (the Monte Carlo estimator exhibits large bias and variance given the large parameter space of BNNs). 

In this work, we  use an alternative way to quantify the model class overlap (we note that this is not a proxy for LML): we still compute Equation $~\ref{eqn:lml}$ but now the datasets $\mathcal{D}$ are drawn from finite-width BNNs or 
the limiting NNGPs while the model $\mathcal{M}$ is the data generating GP. We refer to this metric as log data likelihood (LDL) to avoid confusion with LML. A model has a desirable inductive bias when it can generate diverse datasets with high LDL under the data generating model.

Given a data generating GP with prior $\text{GP}(m(\mathbf{x}),K)$, LDL can be efficiently computed as:
\begin{equation}
\label{eqn:lml_gp}
    \text{LDL} = -\frac{1}{2}(\mathbf{y} - m(\mathbf{x}))^\intercal \Sigma^{-1}_{f} (\mathbf{y} - m(\mathbf{x}))-\frac{1}{2}\log|\Sigma_f|-\frac{n}{2}\log 2\pi, \ \text{  
 where   }\  \Sigma_f = K + \sigma^2_\epsilon I.
\end{equation}

\begin{figure*}[htbp]
     \centering
       \begin{subfigure}[b]{0.49\textwidth}
         \centering
         \includegraphics[width=\textwidth]{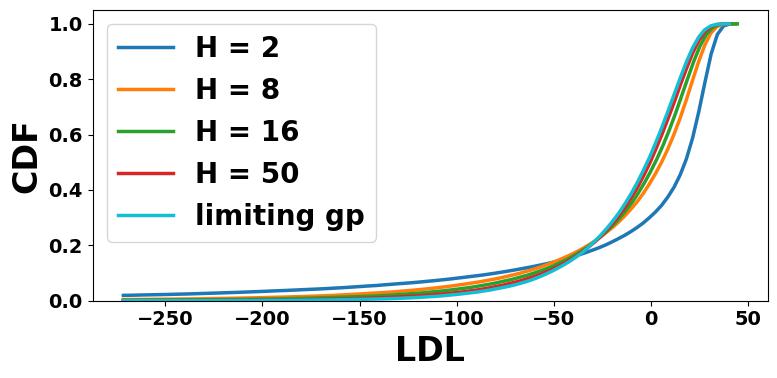}
         \caption{BNN with ReLU, $\sigma_\mathbf{W}^2=\sigma_\mathbf{b}^2=2.0$}
         \label{marginal_log_ll_a}
     \end{subfigure}
     \hfill
    \begin{subfigure}[b]{0.49\textwidth}
         \centering
         \includegraphics[width=\textwidth]{figs/marginal_log_ll/marginal_ll_gp_rbf_2.0_erf_v_2.0.jpg}
         \caption{BNN with erf, $\sigma_\mathbf{W}^2=\sigma_\mathbf{b}^2=2.0$}
         \label{marginal_log_ll_b}
     \end{subfigure}
     \caption{The CDF of LDL (Eqn~\ref{eqn:lml_gp}) with datasets sampled from BNNs with different hidden widths and evaluated under GP-RBF ($l=2.0$): $x$-axis denotes LDL while $y$-axis denotes CDF. Each color represents either a BNN with a specific hidden width or the limiting NNGP. BNNs with smaller widths tend to generate datasets with higher as well as lower LDL than the limiting NNGP.}
\end{figure*}

\textbf{Results} \textbf{Finite- and infinite-width BNNs have distinct inductive biases}. In Figures~\ref{marginal_log_ll_a} and~\ref{marginal_log_ll_b}, we see that BNNs with larger widths generate less diverse datasets: the CDF of larger BNNs is higher in the high LDL region (i.e., they generate limited datasets with high LDL).

\subsection{Qualitatively Comparing Inductive Biases in Function Space} 
\begin{figure}[t]
     \centering
     \begin{subfigure}[b]{0.9\textwidth}
         \centering
         \includegraphics[width=\textwidth]{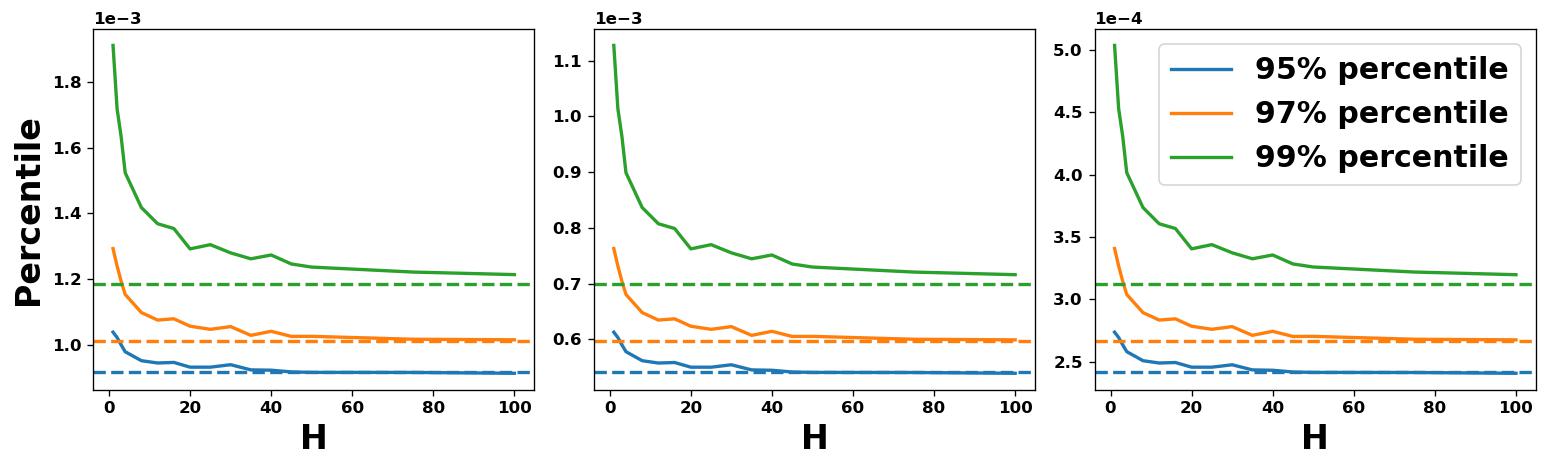}
         \caption{DCT coefficients defined by the prior predictives of BNNs with erf, $\sigma_\mathbf{W}^2=\sigma_\mathbf{b}^2=2.0$}
         \label{fig:dct_arccos_2.0_prior}
     \end{subfigure}
     \hfill
    \begin{subfigure}[b]{0.9\textwidth}
         \centering
         \includegraphics[width=\textwidth]{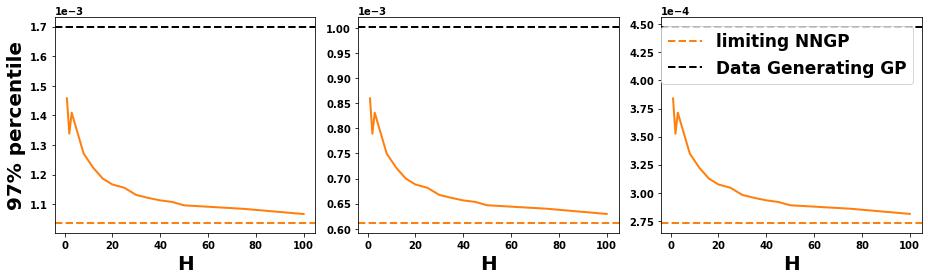}
         \caption{DCT coefficients defined by the posterior predictives of  BNNs with erf, $\sigma_\mathbf{W}^2=\sigma_\mathbf{b}^2=2.0$}
        \label{fig:dct_post_a}
     \end{subfigure}
     \caption{Qualitative analysis of DCT coefficients: Data is generated from GP-RBF ($l=2.0$). Each plot represents a DCT coefficient $a_i$ in vector $\mathbf{a}$.  $x$-axis denotes the hidden width while $y$-axis denotes the $j$-th percentile value of the distribution of $a_i$. Each color represents a different $j$ with the dashed line of the same color representing that of the limiting NNGP. 
     The black dashed line shows the $j$-th percentile of he data generating GP.  A larger percentile implies more mass on large coefficient values.}
\end{figure}
While the LDL allows us to quantitatively compare the inductive biases of two Bayesian model classes, we also want to  qualitatively describe the differences between them. For the latter, we want to compare the differences between the prior and posterior predictive distributions of finite- and infinite-width BNNs in function space. 

Previous works prove that finite-width BNNs yield heavy-tailed priors in function space~\citep{noci2021precise,zavatone2021exact}. However, they do not discuss how heavy-tailed priors affect the generalization performance. 

One common approach to quantifying heavy-tailedness is to compute the higher-order moments of the distribution; however, this is not easy because the function space is large and requires a prohibitively large number of samples. 
We instead analyze the samples drawn from the prior and  posterior predictive distributions using discrete cosine transform (DCT). 

DCT is a linear transformation that expresses a function as a weighted sum of cosines of different frequencies.
Formally, given a function $\mathbf{f}^\intercal = [f_0,\dotsc,f_{N-1}]^\intercal$, DCT is a linear and invertible function, $\mathbf{T}_{\text{DCT}}: \mathbb{R}^N\xrightarrow[]{}\mathbb{R}^N$, that transforms $\mathbf{f}$ as:
\begin{equation}
\label{eqn:dct}
\mathbf{a} = \mathbf{T}_{\text{DCT}}\mathbf{f}, \ \text{ where }\  \mathbf{T}_{\text{DCT}_{ij}} = \frac{\sqrt{2}}{\sqrt{2^{\mathbb{I}[i=0]} N}}\cos\left(\frac{\pi i (2 j + 1)}{2 N} \right),
\end{equation}
where $\mathbf{a}^\intercal=[a_0,\dotsc,a_{N-1}]^\intercal$ are the DCT coefficients representing the weights on cosines of different frequencies in ascending order ($a_0$ is the weight on the lowest frequency and $a_{N-1}$, on the highest).

We qualitatively compare the DCT coefficient distributions of finite- and infinite-width BNNs.

\textbf{Results} 
 Figure~\ref{fig:dct_arccos_2.0_prior} shows that comparing to the NNGPs, \textbf{the functional priors of finite-width BNNs define a spectrum with more mass on large coefficient values across different frequencies} (the $99\%$ percentile of DCT coefficients defined by finite-width BNNs is larger).  
 Figure~\ref{fig:dct_post_a} shows similar trends after inference. Additionally, \textbf{the frequency spectrum defined by finite-width BNNs behaves more similar to that of the data generating model}, which also puts more mass on the larger coefficients (the black dashed line is higher).

\subsection{Connecting the Inductive Bias to Performance Difference Through Frequency Filtering}\label{sec:lowpass_bnn}
In the previous section, we show that the frequency spectrum defined by prior predictive distributions of finite-width BNN places more mass on large coefficient values than that of the limiting NNGP, which makes them more flexible in using functions of different frequencies. In this section, we conduct two sets of experiments to confirm this hypothesis, using ``low-pass filtering''.

\textbf{Low-pass Filtering} Given a function $\mathbf{f}$, we first compute its DCT coefficients $\mathbf{a}$ via Equation~\ref{eqn:dct}, and reconstruct that function via $\mathbf{f} = \mathbf{T}_{\text{DCT}}^\top\mathbf{a}$.
We then remove high-frequency components from the function by setting the corresponding DCT coefficients to $0$: $\tilde{\mathbf{f}} = \mathbf{T}_{\text{\text{DCT}}}^\top \mathbf{I}_{N}^{(k)} \mathbf{T}_{\text{\text{DCT}}} \mathbf{f} := \mathbf{C} \mathbf{f}$,
where $k = tN,\ t\in[0,1]$, and $\mathbf{I}_{N}^{(k)}$ is an $N\times N$ identity matrix with the last $k$ diagonal elements set to $0$.  A larger $t$ removes more high-frequency components, which makes the reconstructed function $\tilde{\mathbf{f}}$ smoother (Figures~\ref{fig:arccos-arccos-qualitative} and \ref{fig:rbf-arccos-qualitative}). $t=0$ recovers the original function. 

\textbf{Low-pass filtered BNNs}
We compare the generalization performance of a finite-width BNN with that of a ``low-pass filtered'' BNN---a BNN in which we filter out high-frequency components from the prior predictive distribution (i.e., we ``smooth'' functions drawn from the functional prior). We note that our approach modifies the function-space prior of BNNs directly, as opposed to post-hoc modifying the posterior predictive distributions, which can be coupled with any inference method (details in Appendix~\ref{apdx:high_freq}).
We find that the model performance decreases now that finite-width BNNs are constrained to the low-frequency regime.  

\begin{figure*}[tbp]
     \centering
     \begin{subfigure}[b]{0.49\textwidth}
         \centering
         \includegraphics[width=\textwidth]{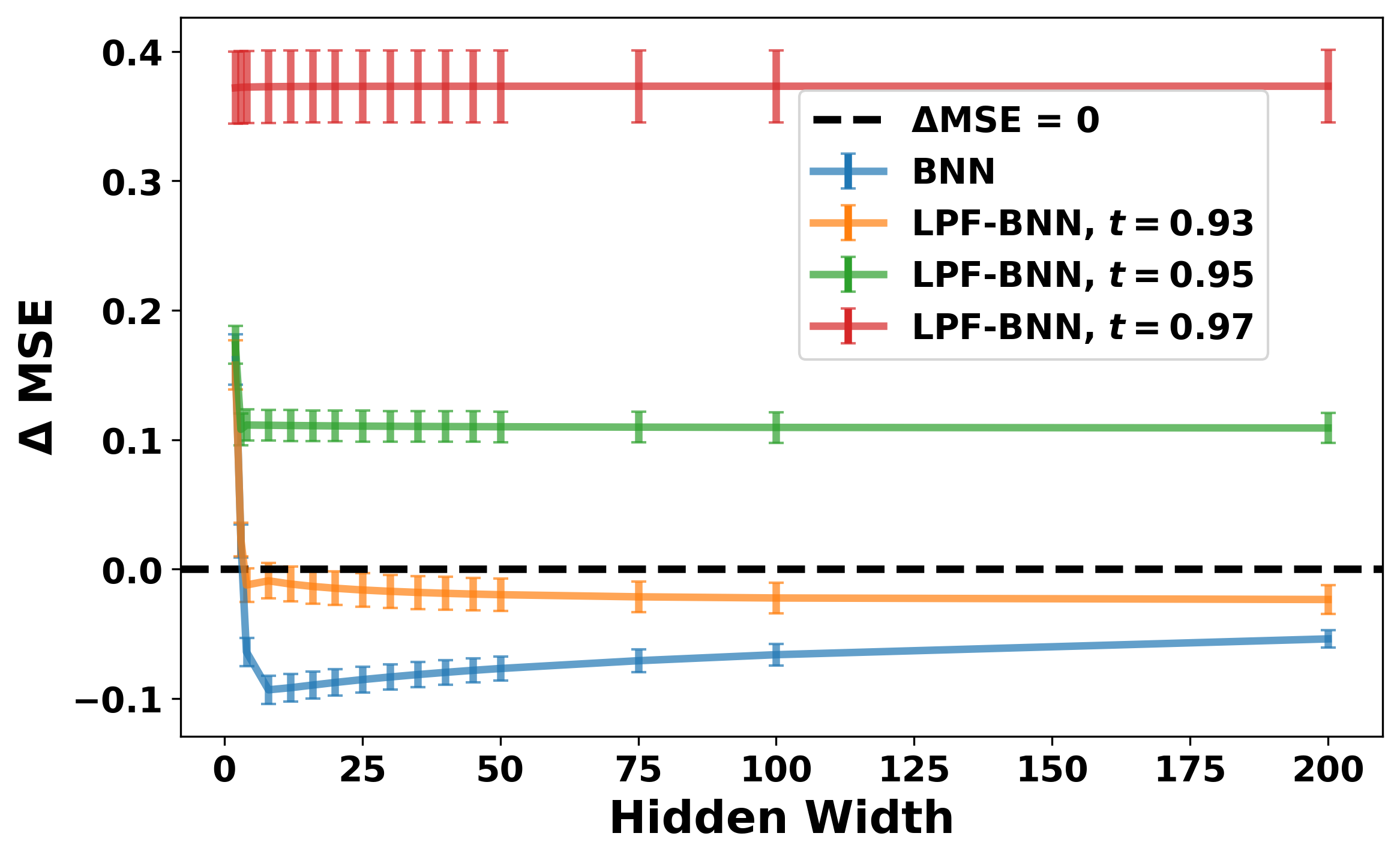}
         \caption{BNN with ReLU, $\sigma_\mathbf{W}^2=\sigma_\mathbf{b}^2=2.0$}
     \end{subfigure}
     \hfill
     \begin{subfigure}[b]{0.49\textwidth}
         \centering
         \includegraphics[width=\textwidth]{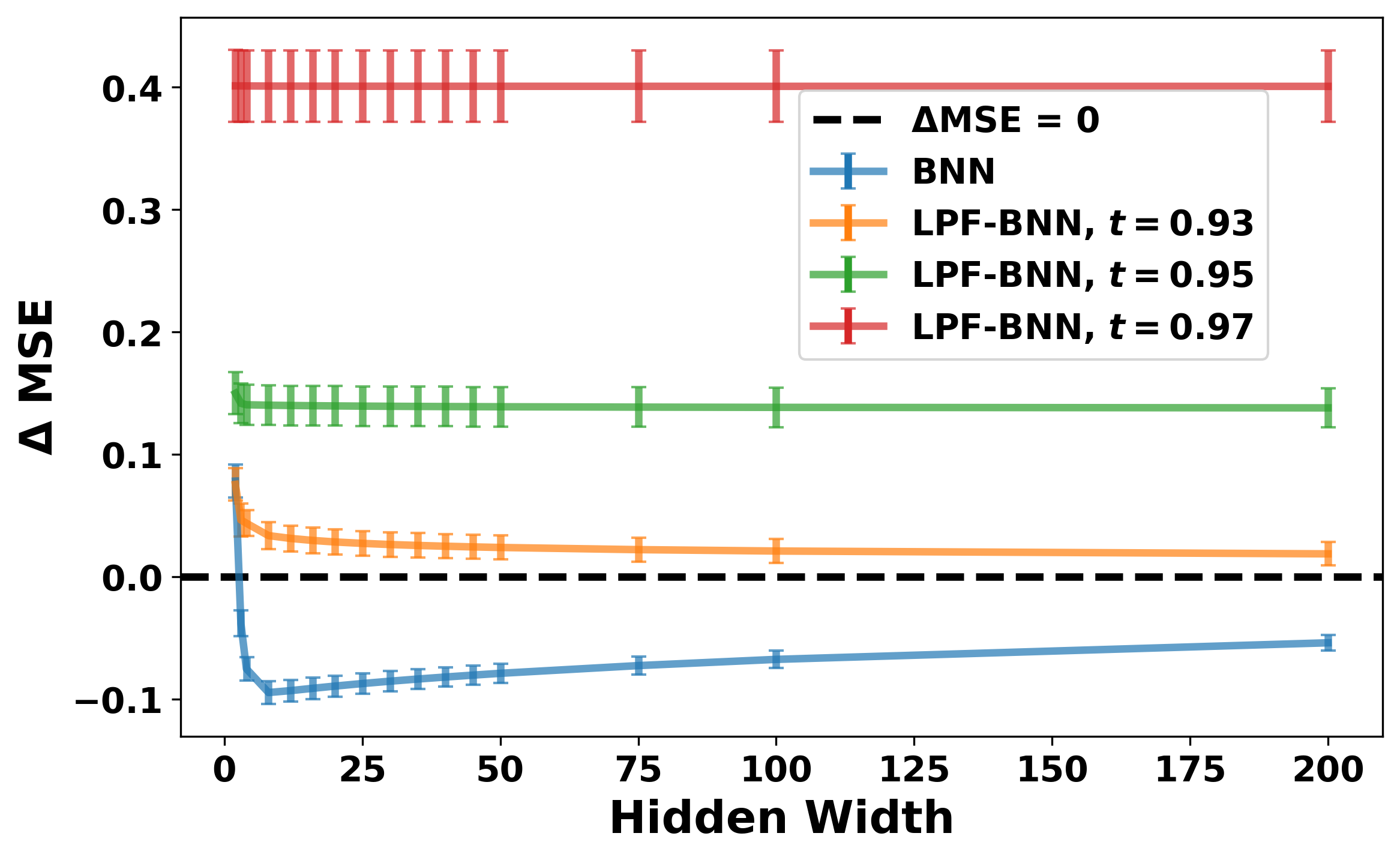}
         \caption{BNN with erf, $\sigma_\mathbf{W}^2=\sigma_\mathbf{b}^2=2.0$}
     \end{subfigure}    
     \caption{Test performance of low-pass filtered BNNs: The data is generated from GP-RBF ($l=0.5$). $x$-axis denotes the hidden-layer width of the BNN while $y$-axis denotes  $\Delta\text{MSE}$.
     Higher thresholds $t$ imply more high frequencies being removed. The generalization performance decreases as $t$ increases. We chose thresholds empirically ($t \leq 0.91$ does not alter the posterior predictive significantly).}
     \label{fig:lpf_bnn_inline}
\end{figure*}

\textbf{Results} \textbf{Removing high-frequency components hurts the generalization performance of finite-width BNNs.} 
In Figure \ref{fig:lpf_bnn_inline}, we show that with more high-frequency components removed ($t$ increases), the generalization performance decreases ($\Delta\text{MSE}$ increases).
Given different BNN architectures, removing the same proportion of the high frequency components has different effects (when $t=0.93$, low-pass filtered BNNs with ReLU outperform the limiting NNGP while low-pass filtered BNNs with erf do not). Additional results are in Appendix \ref{apdx:lpf-bnn-vs-bnn}.

\textbf{Low-pass filtered Datasets}
To provide additional evidence that finite-width BNNs are more flexible, we construct datasets with high-frequency components removed. We expect that finite-width BNNs can adapt to this model-mismatch by putting low mass on high-frequency components. 
We apply low-pass filtering to data generating GPs and compare the DCT spectrum defined by the posterior predictive distributions of finite- and infinite-width BNNs. 

\begin{figure*}[tbp]
     \centering
         \includegraphics[width=\textwidth]{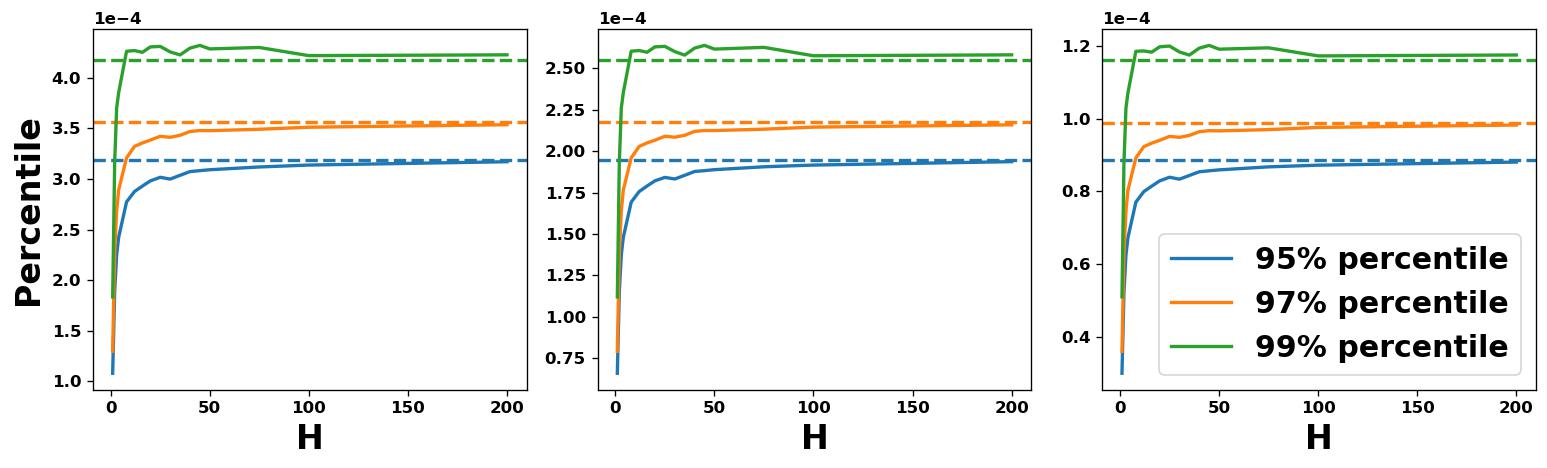}
     \caption{Comparison of DCT coefficients of finite- and infinite-width BNNs: Data is drawn from low-pass filtered GP-Arcsin ($\sigma_\mathbf{W}^2=\sigma_\mathbf{b}^2=2.0$). BNNs with erf, $\sigma_\mathbf{W}^2=\sigma_\mathbf{b}^2=2.0$ are trained. $x$-axis denotes the hidden width while $y$-axis denotes the $j$-th percentile value of the distribution of $a_i$. Each color represents a different $j$ with the dashed line of the same color representing that of the limiting NNGP. 
      A smaller percentile implies more mass around $0$.}
    \label{fig:dct_arcsin_frac}
\end{figure*}

\textbf{Results} 
\textbf{Finite-width BNNs can adapt to the model-mismatch we designed}. Figure ~\ref{fig:dct_arcsin_frac} shows that for finite-width BNNs, the distributions of DCT coefficients of the high-frequency components put more mass around $0$ (i.e., the percentiles of DCT coefficients are lower as the solid line is mostly below the dashed line), which resembles the data generating process ($a_i = 0$ for $i>k$).

\section{Discussion}
In this paper, we quantitatively and qualitatively compare the behaviors of finite- and infinite-width BNNs. We empirically show that finite-width BNNs define a frequency spectrum with more mass on large coefficient values, which allows them to adapt to the data faster under model mismatch. Our experiments suggest that there are advantages to using non-Gaussian models, such as finite-width BNNs or non-Gaussian approximation of the true posterior, even though inference may be challenging. Future work is needed to determine the optimal width given the dataset size.

\section{Acknowledgements}
This material is based upon work supported by the National Science Foundation under Grant No. IIS-2007076.  Any opinions, findings, and conclusions or recommendations expressed in this material are those of the authors and do not necessarily reflect the views of the National Science Foundation.

\bibliographystyle{plainnat}
\bibliography{neurips_2022.bib}


\appendix

\section{Analytical Form for NNGP Kernels}
\label{apdx:sec_nngp_kernel}
Define the following terms,
$$\tilde{\mathbf{x}}^\intercal = [x,\ 1],\ \Sigma = \text{Diag}(\sigma^2_{\mathbf{W}},\sigma^2_{\mathbf{b}}),\ \|\tilde{\mathbf{x}}\|_{\Sigma} = \sqrt{\tilde{\mathbf{x}}^\intercal\Sigma\tilde{\mathbf{x}}}.$$
For a one hidden-layer BNN with erf nonlinearity, the BNN will converge to a GP with Arcsin kernel defined as, 
\begin{equation}\label{apdx:eqn_arcsin}
    K_{\arcsin}(x, x^\prime) = \frac{2\sigma^2_{\mathbf{W}}}{\pi}\theta +  \sigma^2_{\mathbf{b}}\quad
    \text{where}\quad  \theta  = \sin^{-1}\Bigl(\frac{2\tilde{\mathbf{x}}^\intercal \Sigma\tilde{\mathbf{x}}^\prime}{\sqrt{1+2\|\tilde{\mathbf{x}}\|^2_{\Sigma}}\sqrt{1+\|\tilde{\mathbf{x}}^\prime\|^2_{\Sigma}}}\Bigr)
\end{equation}
For a one hidden-layer BNN with ReLU nonlinearity, the BNN will converge to a GP with Arccos kernel defined as,
\begin{equation}\label{apdx:eqn_arccos}
    K_{\arccos}(x, x^\prime) = \frac{\sigma^2_{\mathbf{W}}}{2\pi}\|\tilde{\mathbf{x}}\|_{\Sigma}\|\tilde{\mathbf{x}}^\prime\|_{\Sigma}(\sin \theta + (\pi - \theta)\cos\theta)+\sigma^2_{\mathbf{b}}\quad
  \text{where }\quad\theta  = \cos^{-1}\Bigl(\frac{\tilde{\mathbf{x}}^\intercal \Sigma\tilde{\mathbf{x}}^\prime}{\|\tilde{\mathbf{x}}\|_{\Sigma}\|\tilde{\mathbf{x}}^\prime\|_{\Sigma}}\Bigr)
\end{equation}
The complete derivation can be found in~\citep{williams2006gaussian,cho2011analysis}.

\section{Datasets}\label{apdx:sec_datasets}
\paragraph{Synthetic Datasets}
 We generate data using GPs with three  kernels, RBF, Arcsin and Arccos (GP-Arccos generates rougher functions that require larger model capacity to learn). For GP-RBF, the lengthscale $l$ is chosen from $\{0.5, 1.0, 1.5 ,2.0\}$. For GP-Arcsin/Arccos, the hyperparameters $\sigma^2_{\mathbf{W}}, \sigma^2_{\mathbf{b}}$ in Equation~\ref{apdx:eqn_arcsin},~\ref{apdx:eqn_arccos}, are chosen from $\{0.5, 1.0, 1.5 ,2.0\}$.
For each kernel setting, we draw $200$ functions, each as a univariate dataset. 
For each dataset, we uniformly sample $20$ training points from $[-3,-1]\bigcup[1,3]$ aggregate with $x=0$ to facilitate learning ($N_\text{train} = 21$) and $N_\text{test} = 100$ test point from $[-3,3]$.
We add Gaussian noise with $\sigma_\epsilon = 0.1$ to the training data.
\paragraph{Real Datasets}
We test $3$ real datasets: UCI boston, UCI Energy (the output dimension are fit individually), Sic97~\citep{shah2014student}. We split the data into train, test, valid with the ratio $80-10-10$. We perform cross validation on NNGPs and choose the GP hyper-parameter  with the highest validation MSE.
\section{Models and Inference}\label{apdx:sec_inference}
We fit each dataset using BNNs with erf and ReLU nonlinearity whose prior variances $\sigma^2_{\mathbf{W}}, \sigma^2_{\mathbf{b}}$ are chosen from $\{0.5, 1.0, 1.5 ,2.0\}$ and hidden widths $H$ vary from $2$ to $200$. We use NUTS implemented in NumPyro~\citep{phan2019composable} to obtain posterior samples from $p(\mathbf{W},\mathbf{b}|\mathcal{D})$.

We used $4$ parallel chains, each with $10,000$ samples during burn-in, $40,000$ MCMC samples and a thinning of $5$. 

\section{Low-pass filtered BNNs}
\label{apdx:high_freq}

We modify the original BNN model,
\begin{align}
    \begin{split}
        \mathbf{W}, \mathbf{b} &\sim p(\mathbf{W}, \mathbf{b}) \\   
        y | \mathbf{x}, \mathbf{W}, \mathbf{b} &\sim \mathcal{N}(y | f_\text{NN}(\mathbf{x}; \mathbf{W}, \mathbf{b}), \sigma^2_\epsilon),
    \end{split}
    \label{eq:pre-lpf-bnn}
\end{align}
so that the $f_\text{NN}$ is filtered to include only its low-frequency components. We denote this low-pass filtered function by $\tilde{f}_\text{NN}$, 
and describe a general algorithm for computing $\tilde{f}_\text{NN}$ and incorporating it into the model, allowing it to be paired with \emph{any} inference algorithm (e.g., NUTS, which requires $\tilde{f}_\text{NN}$ to be differentiable). 
For simplicity, we assume the inputs are 1-dimensional ($d_{\text{in}} = 1$). 

\paragraph{Step 1: From weight-space to function space.}
To filter out high-frequency components of a \emph{function} $f_\text{NN}$, we cannot simply operate on the corresponding neural network \emph{weights} $\mathbf{W}, \mathbf{b}$. Instead, we decompose the function (evaluated on a grid of points) into a weighted sum of cosines (using the DCT) and then remove the highest frequency terms in the sum. 
Specifically, we define a grid of $N^\text{grid}$ uniformly spaced inputs $\mathbf{x}^\text{grid}_n \in \mathbb{R}^{d_{\text{in}}}$,
denoted jointly by $\mathbf{X}^\text{grid} \in \mathbb{R}^{N^\text{grid} \times d_{\text{in}}}$.
We perform the DCT on $f_\text{NN}(\cdot; \mathbf{W}, \mathbf{b})$ evaluated on this grid.
We further ensure that our training and test sets lie on this grid by mapping them to their nearest neighbors on the grid (a fine grid will only negligibly modify the training and test sets).

\paragraph{Step 2: Performing the DCT.}
We compute the type-II orthonormal DCT, $\mathbf{a}_k(\mathbf{W}, \mathbf{b}) \in \mathbb{R}^{N^\text{grid}}$, of $f_\text{NN}(\mathbf{X}^\text{grid}; \mathbf{W}, \mathbf{b})$ as follows:
\begin{equation}
    \mathbf{a}_k(\mathbf{W}, \mathbf{b}) = 
    \frac{\sqrt{2}}{\sqrt{2^{\mathbb{I}[k = 0]} N^\text{grid} }} \cdot
    \sum_{n=0}^{N^\text{grid}-1} f_\text{NN}(\mathbf{x}_n^\text{grid}; \mathbf{W}, \mathbf{b}) \cdot \cos\left(\frac{\pi k (2n+1)}{2 N^\text{grid}} \right),
    \end{equation}
for $k=0,\dotsc,N^\text{grid}-1$.

\paragraph{Step 3: Low-pass Filtering.}
We select a filtering threshold $t \in [0,1]$, where $t = 0$ recovers the original $f_\text{NN}$, and as $t$ increases, frequencies are filtered out from highest to lowest. 
We then zero out all DCT coefficients $\mathbf{a}_k(\mathbf{W}, \mathbf{b})$ for which $k \geq (1-t) N^\text{grid}$, and construct the low-pass filtered function using the type-II inverse orthonormal discrete cosine transformation:
\begin{align}
\begin{split}
\tilde{f}_\text{NN}(\mathbf{x}_n^\text{grid}; \mathbf{W}, \mathbf{b}) = &\frac{1}{\sqrt{N^\text{grid}}} \cdot \mathbb{I}[t < 1] \cdot \mathbf{a}_0(\mathbf{W}, \mathbf{b}) \\
&+ \frac{\sqrt{2}}{\sqrt{N^\text{grid}}} \cdot \sum_{k=1}^{N^\text{grid}-1} \mathbb{I}[k < (1-t) N^\text{grid}] \cdot \mathbf{a}_n(\mathbf{W}, \mathbf{b}) \cdot \cos\left(\frac{\pi k (2n+1)}{2 N^\text{grid}} \right),
\end{split}
\end{align}
for $n=0,\dotsc,N^\text{grid}-1$. 

\paragraph{Step 4: Inference.}
Lastly, we incorporate the low-pass filtered function into the BNN model by replacing $f_\text{NN}$ with $\tilde{f}_\text{NN}$ in the model in Equation \ref{eq:pre-lpf-bnn} as follows:
\begin{align}
    \begin{split}
        \mathbf{W}, \mathbf{b} &\sim p(\mathbf{W}, \mathbf{b}) \\   
        y | \mathbf{x}, \mathbf{W}, \mathbf{b} &\sim \mathcal{N}(y | \mathbf{0}_\mathbf{x}^\intercal \cdot \tilde{f}_\text{NN}(\mathbf{X}^\text{grid}; \mathbf{W}, \mathbf{b}), \sigma^2_\epsilon),
    \end{split}
\end{align}
where $\mathbf{0}_\mathbf{x}$ is a vector of 0s with a 1 corresponding to the row of $\mathbf{X}^\text{grid}$ closest to $\mathbf{x}$.
Since the low-pass filter is a linear operation, it is differentiable and enables pairing with any inference method available in standard probabilistic programming languages. 

\paragraph{Computational Efficiency.} We can vectorize the above algorithm as follows:
\begin{enumerate}
    \item Let $\mathbf{T} \in \mathbb{R}^{N^\text{grid} \times N^\text{grid}}$ denote the matrix corresponding to the type-II orthonormal discrete cosine transform on a grid of points $\mathbf{X}^\text{grid} \in \mathbb{R}^{N^\text{grid}}$: 
    \begin{equation}
        \mathbf{T}_{\text{DCT}_{kn}} = \frac{\sqrt{2}}{\sqrt{2^{\mathbb{I}[k=0]} N^\text{grid}}}\cos\left(\frac{\pi k (2 n + 1)}{2 N^\text{grid}} \right)
    \end{equation}
    So, $\mathbf{a} = \mathbf{T}_{\text{DCT}} \mathbf{f}$ and $\mathbf{T}_{\text{DCT}}^\top$ is the inverse transform. 
    \item Let $\mathbf{I}_{N^\text{grid}, t}$ denote an identity matrix of size $N^\text{grid}$ with $(\mathbf{I}_{N^\text{grid}, t})_{kk} = 0$ for $k \geq (1-t) N^\text{grid}$. 
    \item Let $\tilde{f}_\text{NN}(\mathbf{X}^\text{grid}; \mathbf{W}, \mathbf{b}) = \mathbf{T}_{\text{DCT}}^\top \mathbf{I}_{N^\text{grid}, t} \mathbf{T}_{\text{DCT}} \cdot f_\text{NN}(\mathbf{X}^\text{grid}; \mathbf{W}, \mathbf{b})$.
    \item For a batch of inputs $\mathbf{X}\in\mathbb{R}^{N\times N^\text{grid}}$, we can write the likelihood as 
    \begin{equation}
        p(\mathbf{y} | \mathbf{X},\mathbf{W}) = \mathcal{N}(\mathbf{y} | \mathbf{0}_\mathbf{X} \tilde{f}_\text{NN}(\mathbf{X}^\text{grid}; \mathbf{W}, \mathbf{b}),\sigma^2_\epsilon),
    \end{equation}
    where $\mathbf{0}_\mathbf{X} = (\mathbf{0}_{\mathbf{x}_1}^\intercal, \dotsc, \mathbf{0}_{\mathbf{x}_N}^\intercal)^\intercal \in \mathbb{R}^{N\times N^\text{grid}}$ is a matrix of 0s and a single 1 in each row that functions to select the elements of $\tilde{f}_\text{NN}(\mathbf{X}^\text{grid}; \mathbf{W}, \mathbf{b})$ corresponding to the training inputs $\mathbf{x}_1,\dotsc,\mathbf{x}_N$.
\end{enumerate}

\FloatBarrier

\section{Additional Results}

\subsection{Generalization Performance of Finite and Infinite Width BNNs}\label{apdx:subsec_data_fit}
\begin{figure*}[htbp]
     \centering
     \begin{subfigure}[b]{0.49\textwidth}
         \centering
         \includegraphics[width=\textwidth]{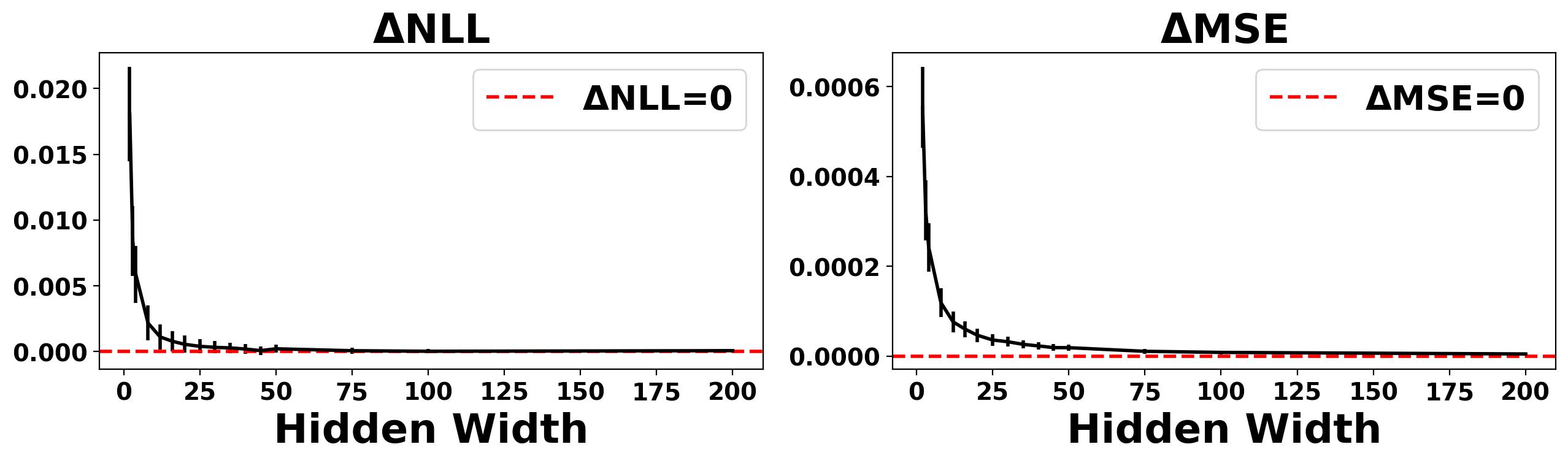}
         \caption{BNN with ReLU, $\sigma_\mathbf{W}^2=\sigma_\mathbf{b}^2=0.5$}
         \label{arccos_0.5_relu_0.5}
     \end{subfigure}
      \hfill
     \begin{subfigure}[b]{0.49\textwidth}
         \centering
         \includegraphics[width=\textwidth]{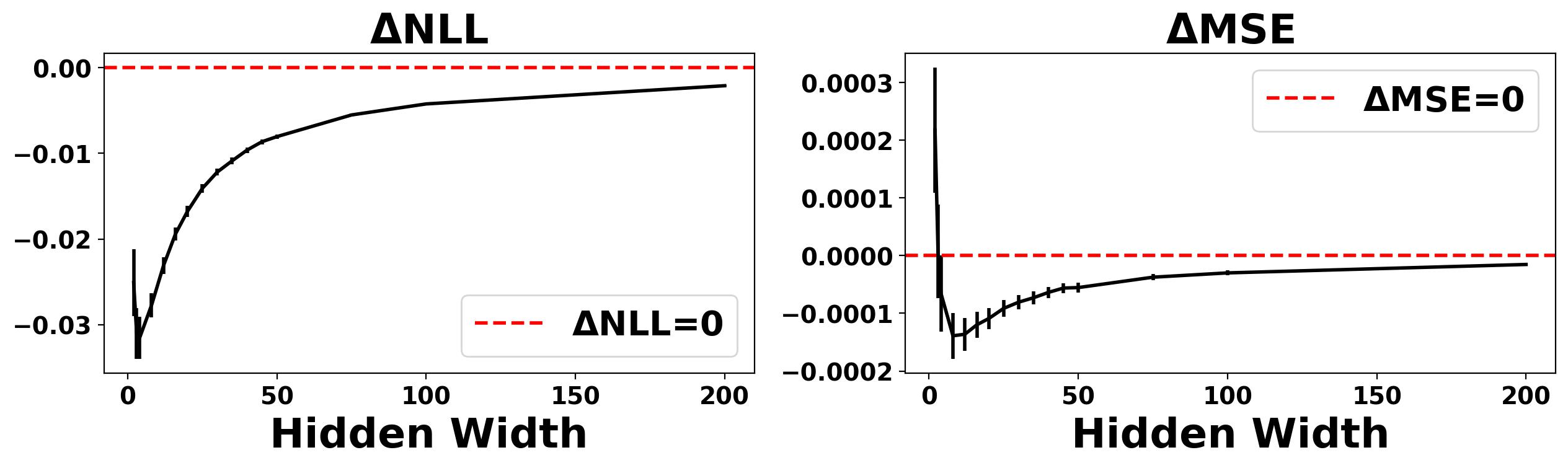}
         \caption{BNN with ReLU, $\sigma_\mathbf{W}^2=\sigma_\mathbf{b}^2=1.0$}
         \label{arccos_0.5_relu_1.0}
     \end{subfigure}
     \hfill
          \begin{subfigure}[b]{0.49\textwidth}
         \centering
         \includegraphics[width=\textwidth]{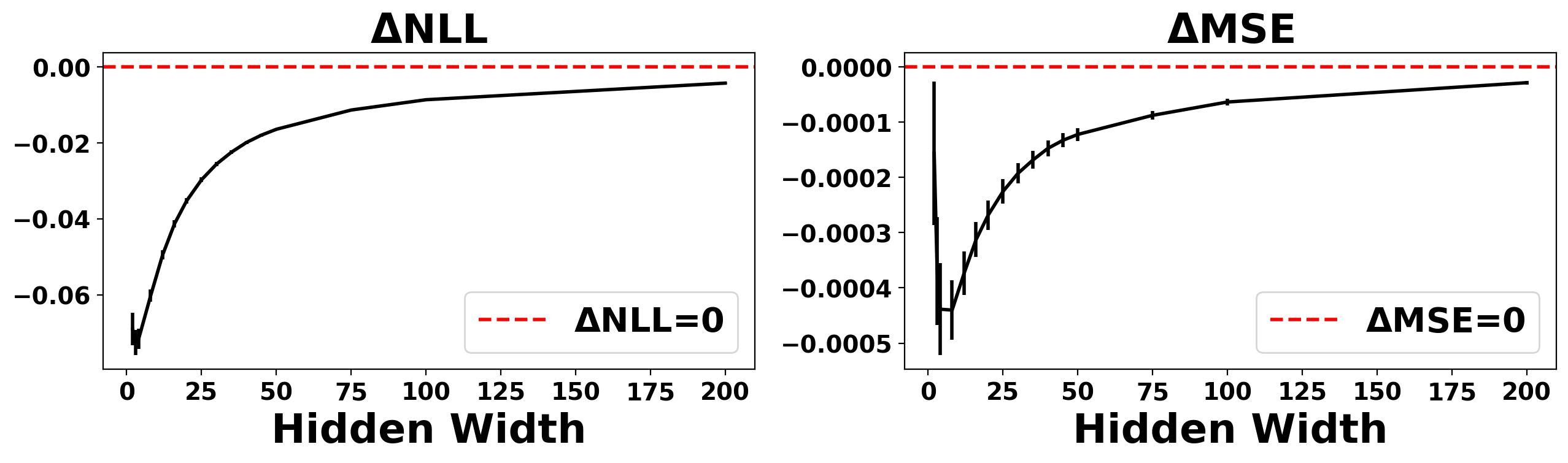}
         \caption{BNN with ReLU, $\sigma_\mathbf{W}^2=\sigma_\mathbf{b}^2=1.5$}
         \label{arccos_0.5_relu_1.5}
     \end{subfigure}
    \hfill
     \begin{subfigure}[b]{0.49\textwidth}
         \centering
         \includegraphics[width=\textwidth]{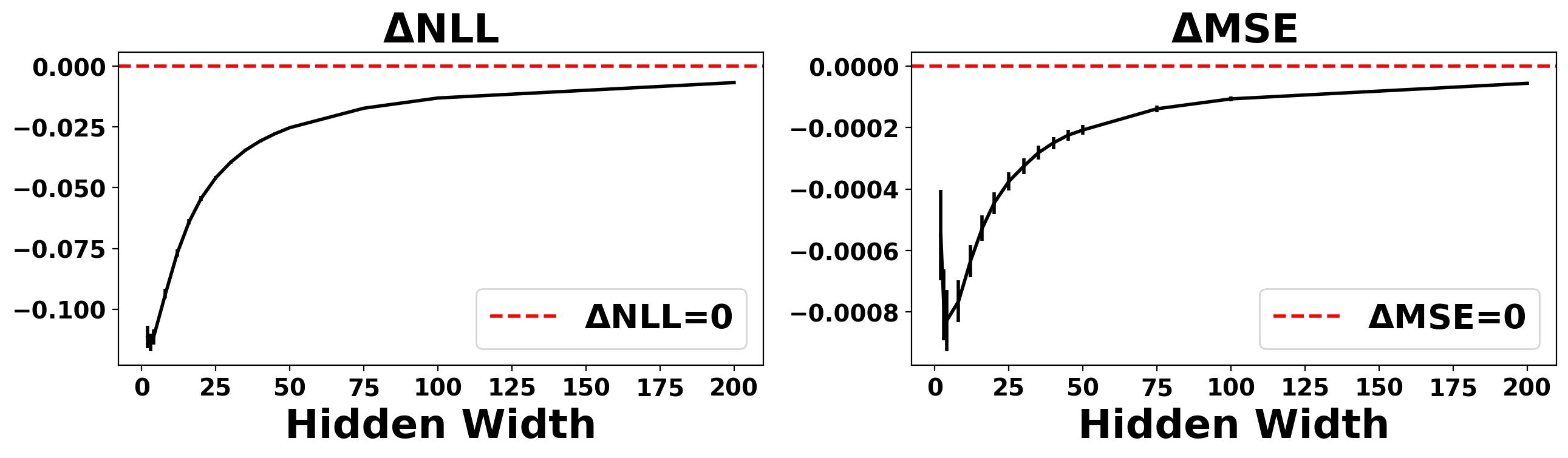}
         \caption{BNN with ReLU, $\sigma_\mathbf{W}^2=\sigma_\mathbf{b}^2=2.0$}
         \label{arccos_0.5_relu_2.0}
     \end{subfigure}
     \caption{Plots of the test performance of finite width BNNs:  The data is generated from GP-Arccos with $\sigma_\mathbf{W}^2=\sigma_\mathbf{b}^2=2.0$ and fit with BNNs with ReLU. There is no model mismatch in Figure~\ref{arccos_0.5_relu_0.5}. The performance gap increases as the model mismatch increases (From Figure~\ref{arccos_0.5_relu_1.0}) to~\ref{arccos_0.5_relu_2.0}, the lowest point of the black curves gets smaller.}
     \label{fig:appdx_arccos_0.5}
\end{figure*}

\begin{figure*}[htbp]
     \centering
     \begin{subfigure}[b]{0.49\textwidth}
         \centering
         \includegraphics[width=\textwidth]{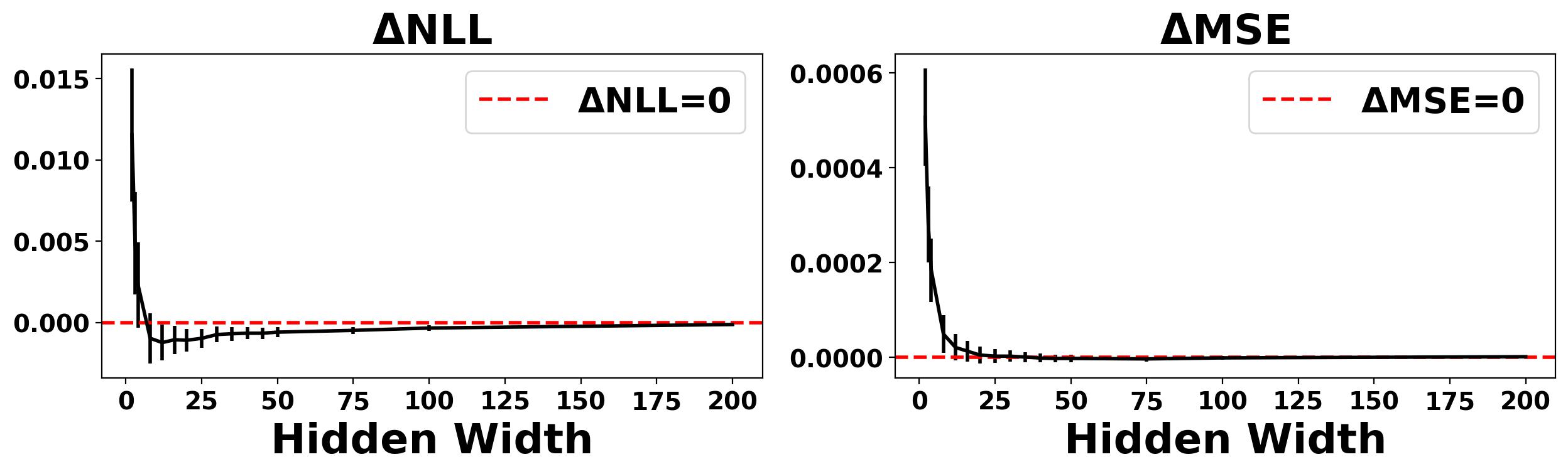}
         \caption{BNN with erf, $\sigma_\mathbf{W}^2=\sigma_\mathbf{b}^2=0.5$}
         \label{arcsin_0.5_erf_0.5}
     \end{subfigure}
      \hfill
     \begin{subfigure}[b]{0.49\textwidth}
         \centering
         \includegraphics[width=\textwidth]{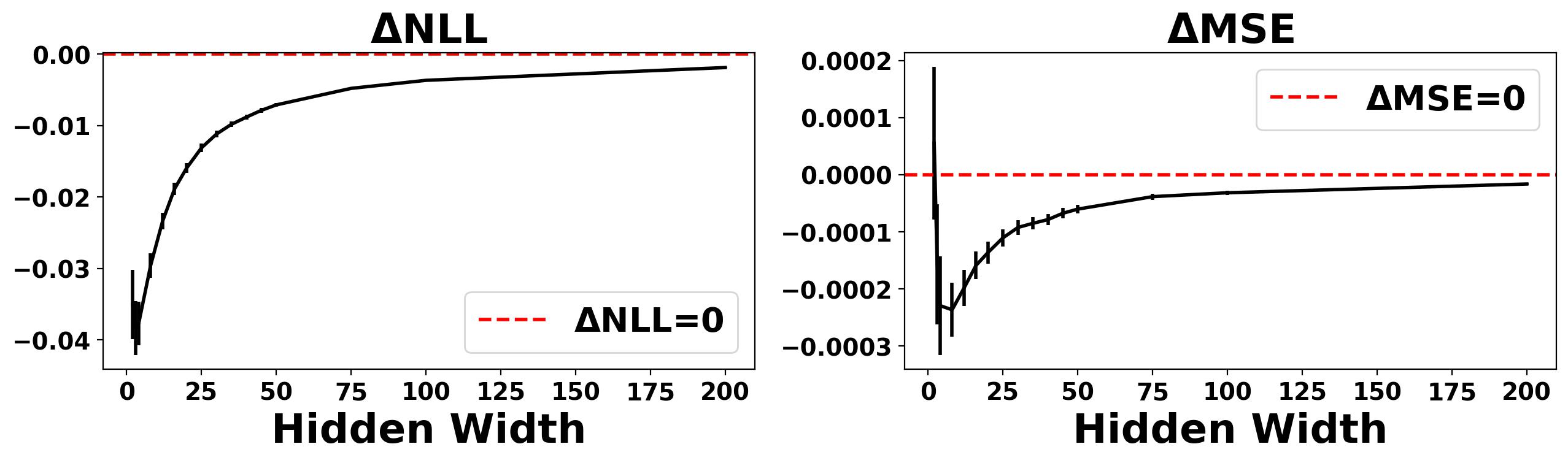}
         \caption{BNN with erf, $\sigma_\mathbf{W}^2=\sigma_\mathbf{b}^2=1.0$}
         \label{arcsin_0.5_erf_1.0}
     \end{subfigure}
     \hfill
          \begin{subfigure}[b]{0.49\textwidth}
         \centering
         \includegraphics[width=\textwidth]{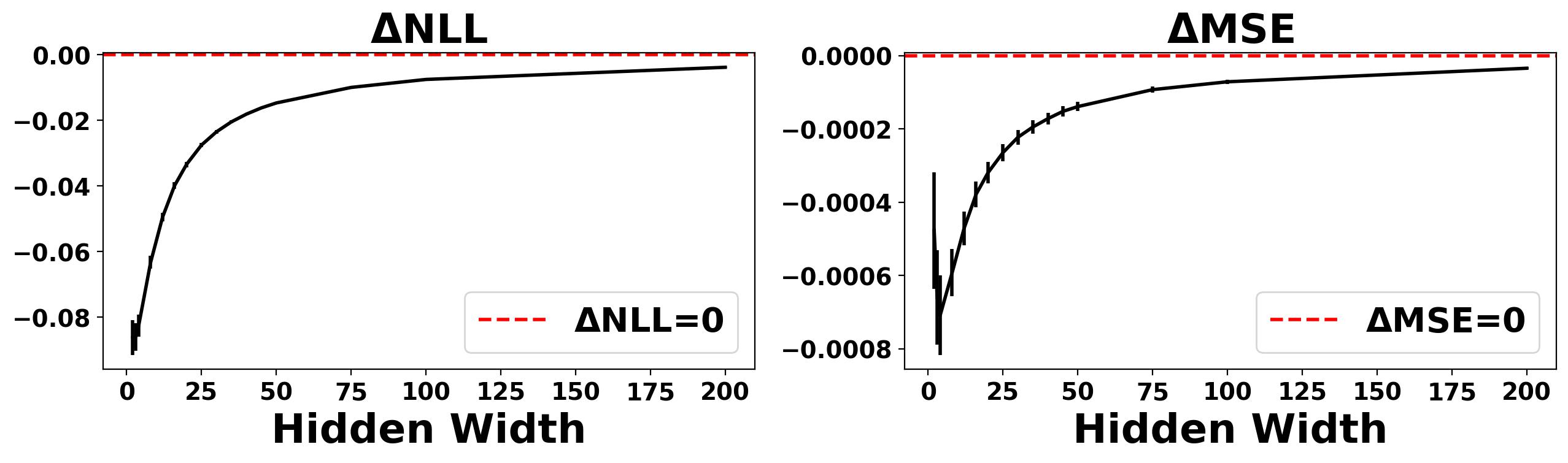}
         \caption{BNN with erf, $\sigma_\mathbf{W}^2=\sigma_\mathbf{b}^2=1.5$}
         \label{arcsin_0.5_erf_1.5}
     \end{subfigure}
    \hfill
     \begin{subfigure}[b]{0.49\textwidth}
         \centering
         \includegraphics[width=\textwidth]{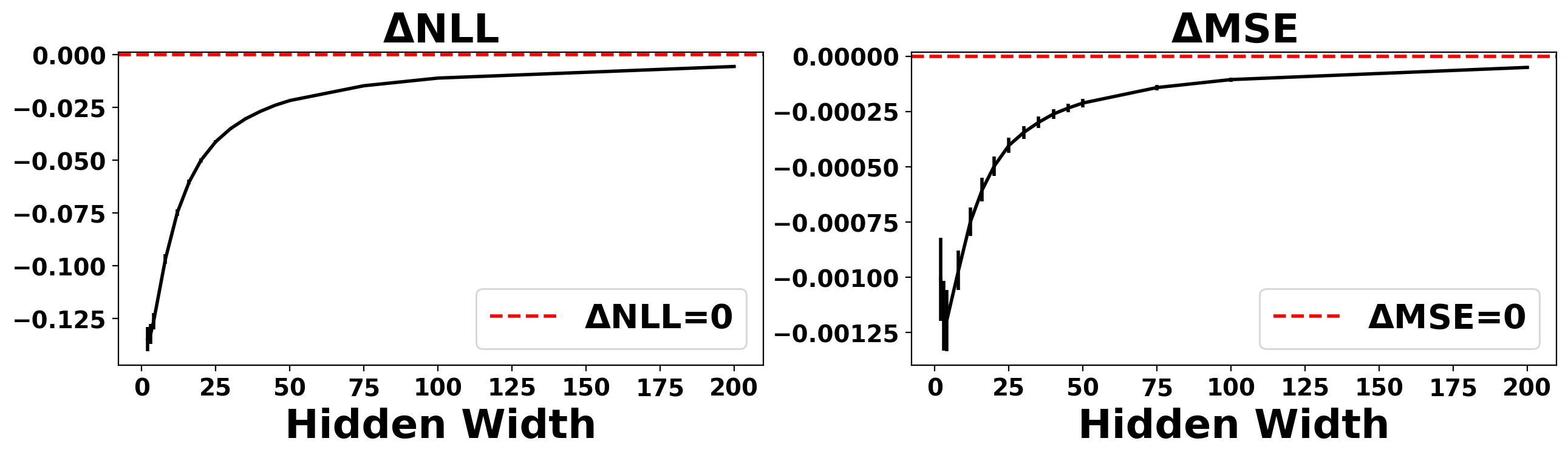}
         \caption{BNN with erf, $\sigma_\mathbf{W}^2=\sigma_\mathbf{b}^2=2.0$}
         \label{arcsin_0.5_erf_2.0}
     \end{subfigure}
     \caption{Plots of the test performance of finite width BNNs:  The data is generated from GP-Arcsin with $\sigma_\mathbf{W}^2=\sigma_\mathbf{b}^2=2.0$ and fit with BNNs with erf. There is no model mismatch in Figure~\ref{arcsin_0.5_erf_0.5}. The performance gap increases as the model mismatch increases (From Figure~\ref{arcsin_0.5_erf_1.0}) to~\ref{arcsin_0.5_erf_2.0}, the lowest point of the black curves gets smaller.}
     \label{fig:appdx_arcsin_0.5}
\end{figure*}

\begin{figure*}[htbp]
     \centering
     \begin{subfigure}[b]{0.49\textwidth}
         \centering
         \includegraphics[width=\textwidth]{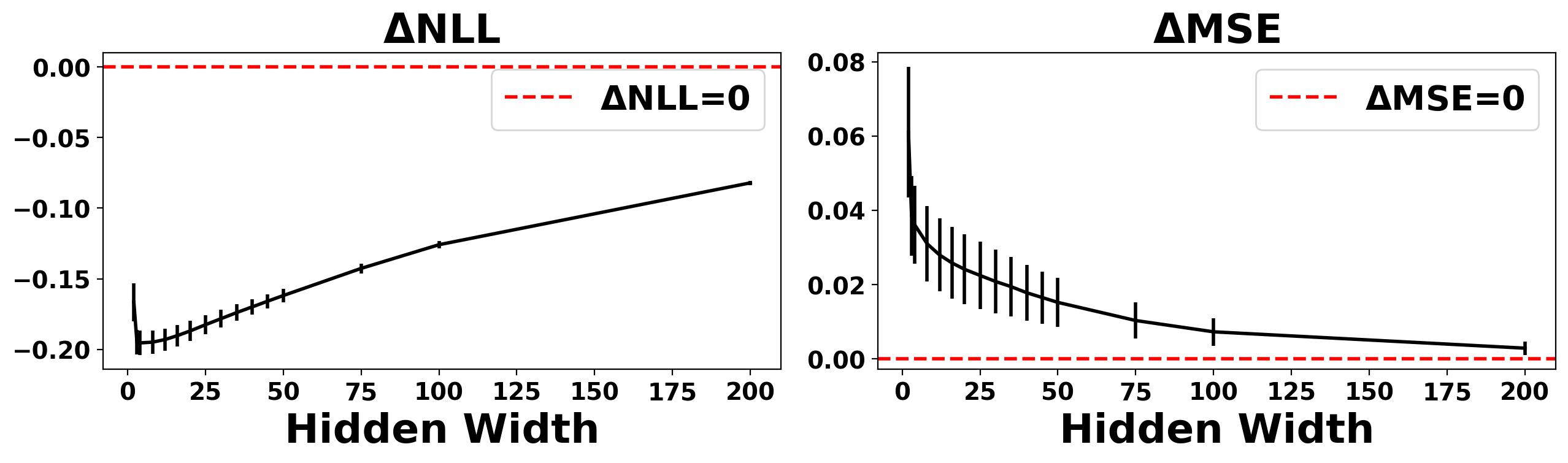}
         \caption{UCI Boston: BNNs with ReLU, $\sigma_\mathbf{W}^2=\sigma_\mathbf{b}^2=3.0,\ \sigma_\epsilon=0.5$}
         \label{uci_boston}
     \end{subfigure}
      \hfill
     \begin{subfigure}[b]{0.49\textwidth}
         \centering
         \includegraphics[width=\textwidth]{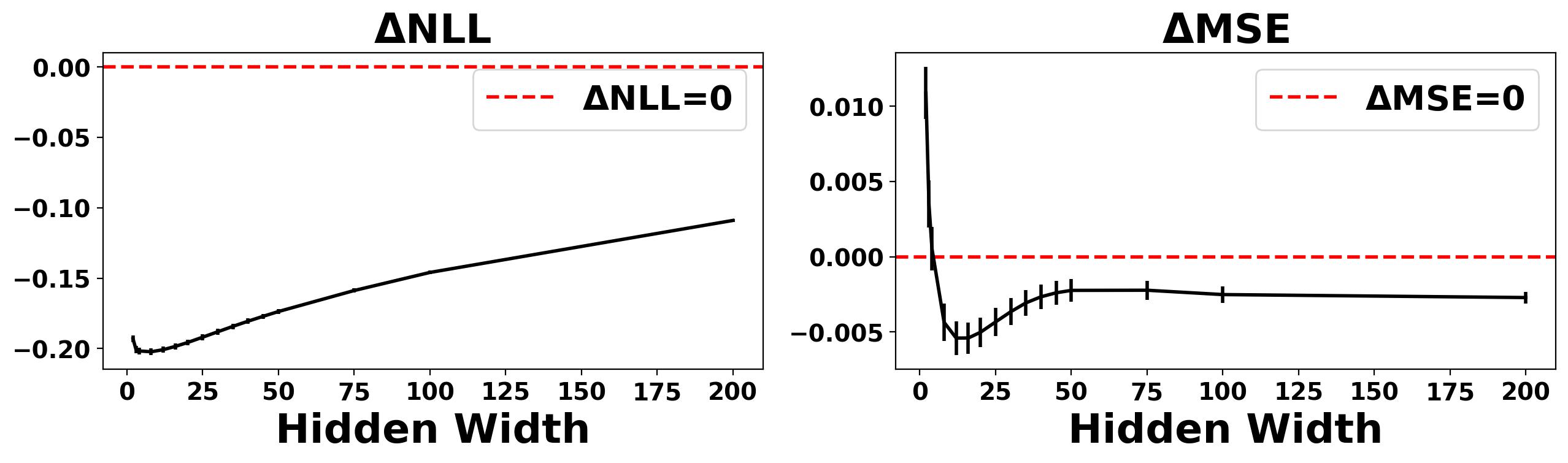}
        \caption{UCI Energy (first output dimension): BNNs with ReLU, $\sigma_\mathbf{W}^2=\sigma_\mathbf{b}^2=5.0,\ \sigma_\epsilon=0.5$}
         \label{uci_energ_d1}
     \end{subfigure}
       \hfill
     \begin{subfigure}[b]{0.49\textwidth}
         \centering
         \includegraphics[width=\textwidth]{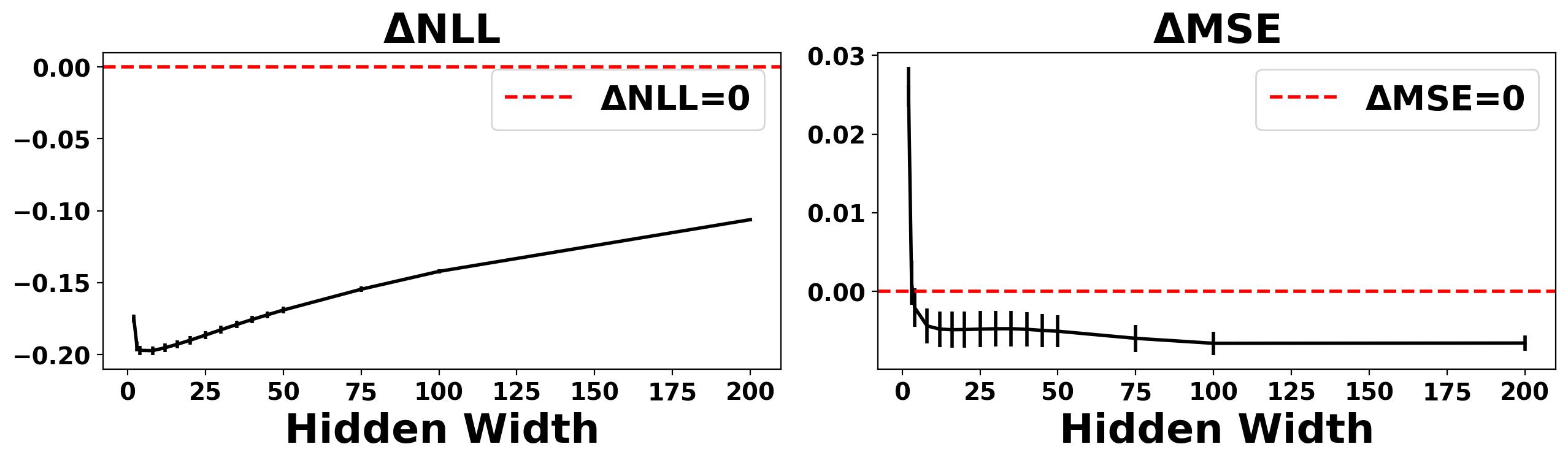}
        \caption{UCI Energy (second output dimension): BNNs with ReLU, $\sigma_\mathbf{W}^2=\sigma_\mathbf{b}^2=5.0,\ \sigma_\epsilon=0.5$}
         \label{uci_energ_d2}
     \end{subfigure}
     \hfill
    \begin{subfigure}[b]{0.49\textwidth}
         \centering
         \includegraphics[width=\textwidth]{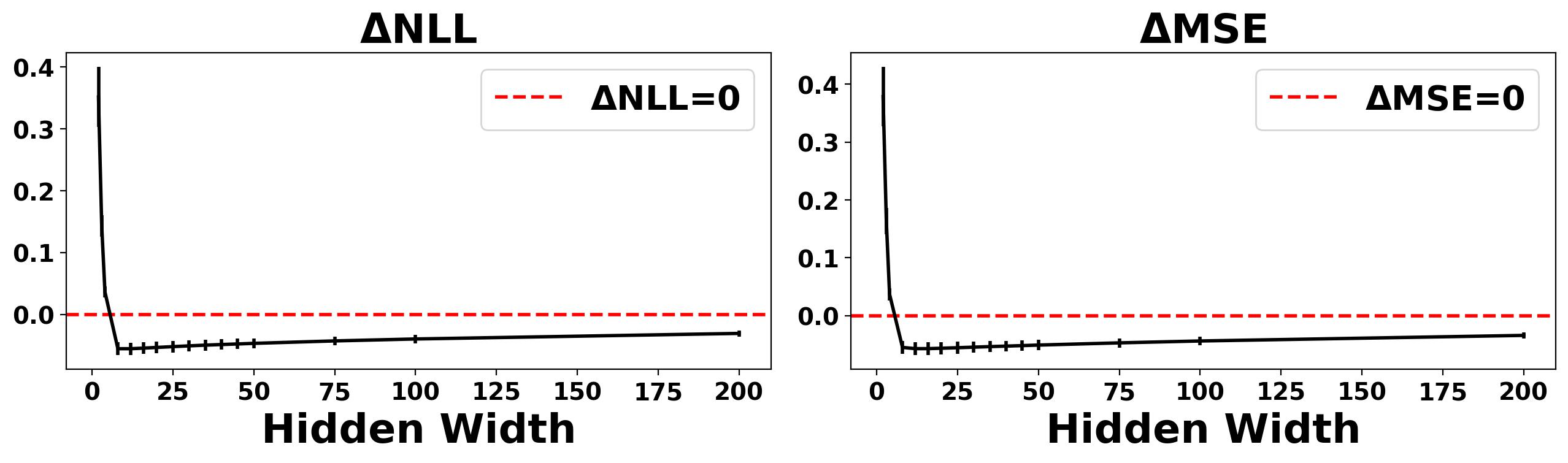}
        \caption{Sic-97~\citep{shah2014student}: BNNs with ReLU, $\sigma_\mathbf{W}^2=\sigma_\mathbf{b}^2=5.0,\ \sigma_\epsilon=0.5$}
         \label{uci_sic97}
     \end{subfigure}
     \caption{Plots of the test performance of finite width BNNs on real datasets: There is often model mismatch in real life. We see that except for UCI Boston, finite width BNNs have better generalization performance.}
     \label{fig:uci}
\end{figure*}

\FloatBarrier
We show additional results comparing the generalization performance of finite and infinite width BNNs. Under all model mismatch (the limiting NNGP is not equivalent to the data generating model) we design, finite width BNNs can outperform infinite width BNNs. Additionally, we observe that larger model mismatch results in larger performance gap (Figure~\ref{fig:appdx_arccos_0.5},~\ref{fig:appdx_arcsin_0.5}). We also provide empirical results on the real dataset (Figure~\ref{fig:uci}) and we see that finite width BNNs perform better on large datasets (Figure~\ref{uci_energ_d1},~\ref{uci_energ_d2},~\ref{uci_sic97}).
\begin{figure*}[htbp]
     \centering
     \begin{subfigure}[b]{0.49\textwidth}
         \centering
         \includegraphics[width=\textwidth]{figs/data_fit/data_rbf_v_0.5_relu_v_2.0_trial_0.jpg}
         \caption{BNN with ReLU, $\sigma_\mathbf{W}^2=\sigma_\mathbf{b}^2=0.5$}
         \label{rbf_0.5_relu_0.5}
     \end{subfigure}
      \hfill
     \begin{subfigure}[b]{0.49\textwidth}
         \centering
         \includegraphics[width=\textwidth]{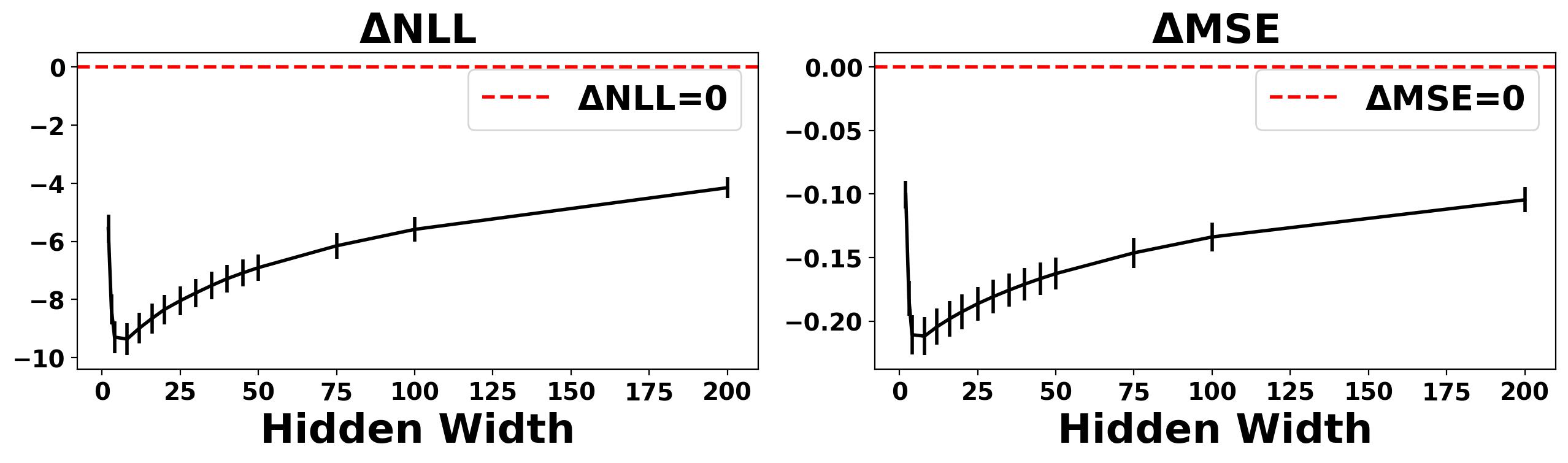}
         \caption{BNN with erf, $\sigma_\mathbf{W}^2=\sigma_\mathbf{b}^2=0.5$}
         \label{rbf_0.5_erf_0.5}
     \end{subfigure}
     \hfill
          \begin{subfigure}[b]{0.49\textwidth}
         \centering
         \includegraphics[width=\textwidth]{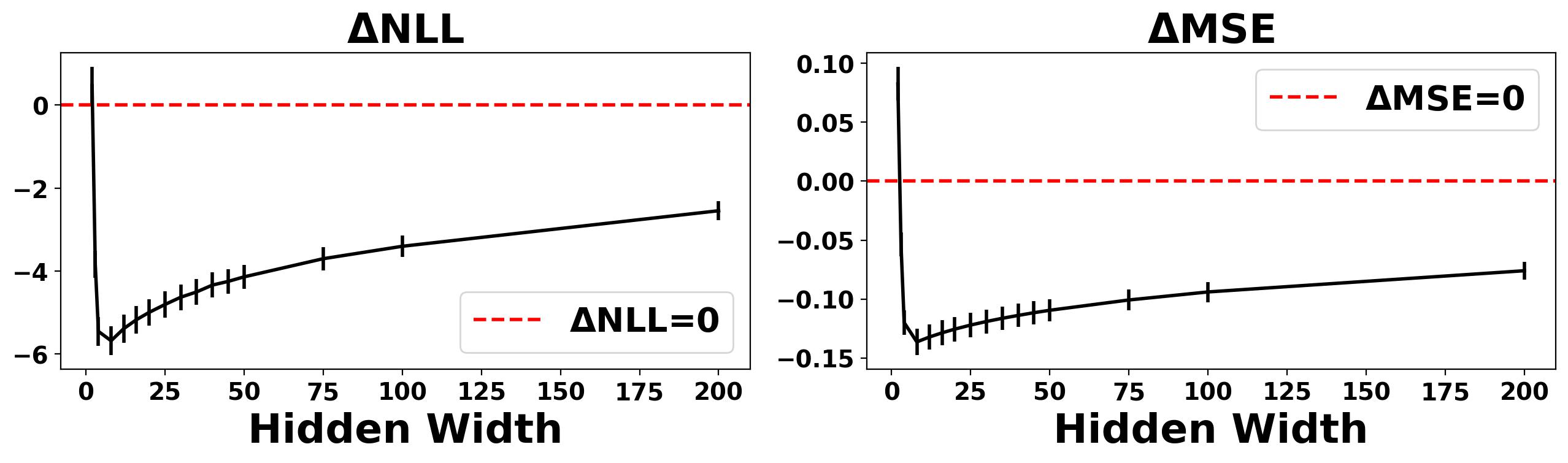}
         \caption{BNN with ReLU, $\sigma_\mathbf{W}^2=\sigma_\mathbf{b}^2=1.0$}
         \label{rbf_0.5_relu_1.0}
     \end{subfigure}
     \hfill
     \begin{subfigure}[b]{0.49\textwidth}
         \centering
         \includegraphics[width=\textwidth]{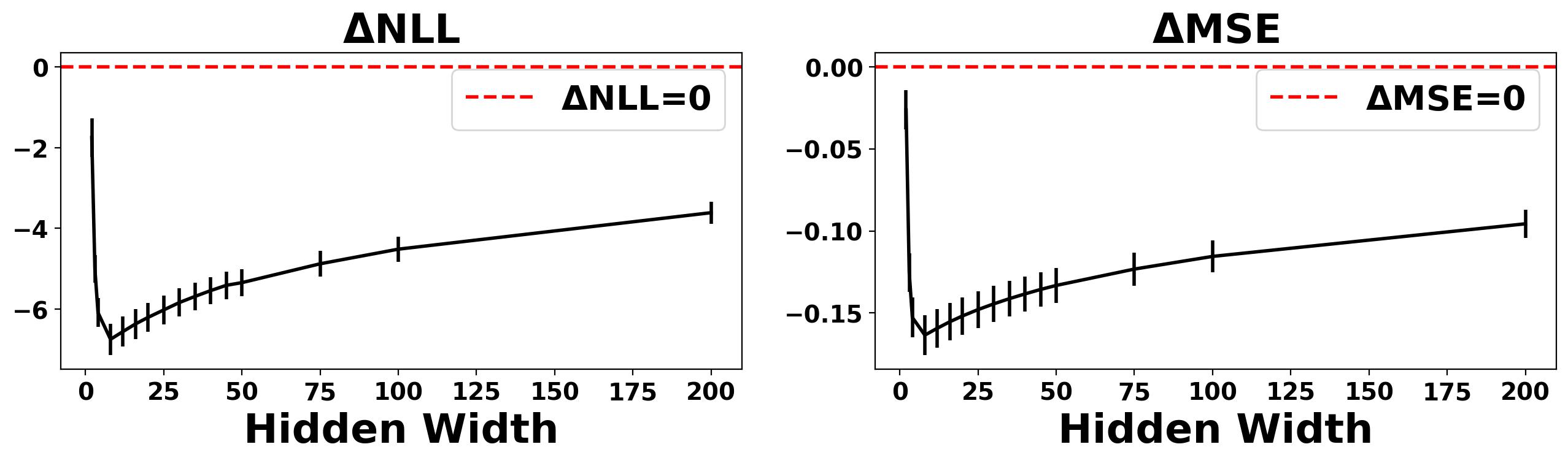}
         \caption{BNN with erf, $\sigma_\mathbf{W}^2=\sigma_\mathbf{b}^2=1.0$}
         \label{rbf_0.5_erf_1.0}
     \end{subfigure}
    \hfill
     \begin{subfigure}[b]{0.49\textwidth}
         \centering
         \includegraphics[width=\textwidth]{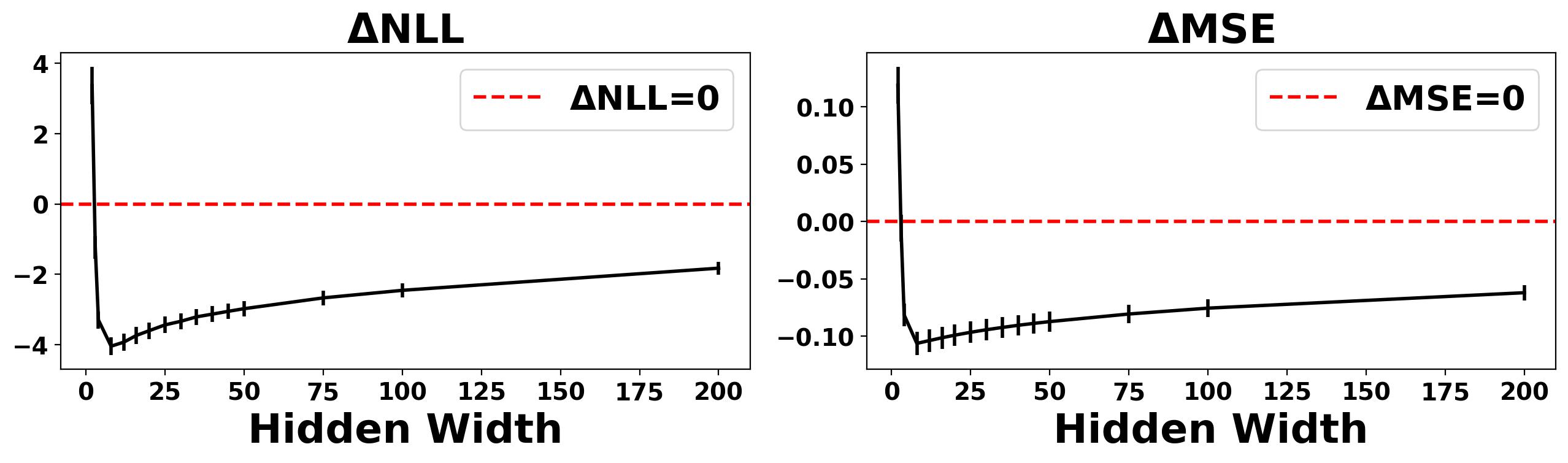}
         \caption{BNN with ReLU, $\sigma_\mathbf{W}^2=\sigma_\mathbf{b}^2=1.5$}
         \label{rbf_0.5_relu_1.5}
     \end{subfigure}
     \hfill
     \begin{subfigure}[b]{0.49\textwidth}
         \centering
         \includegraphics[width=\textwidth]{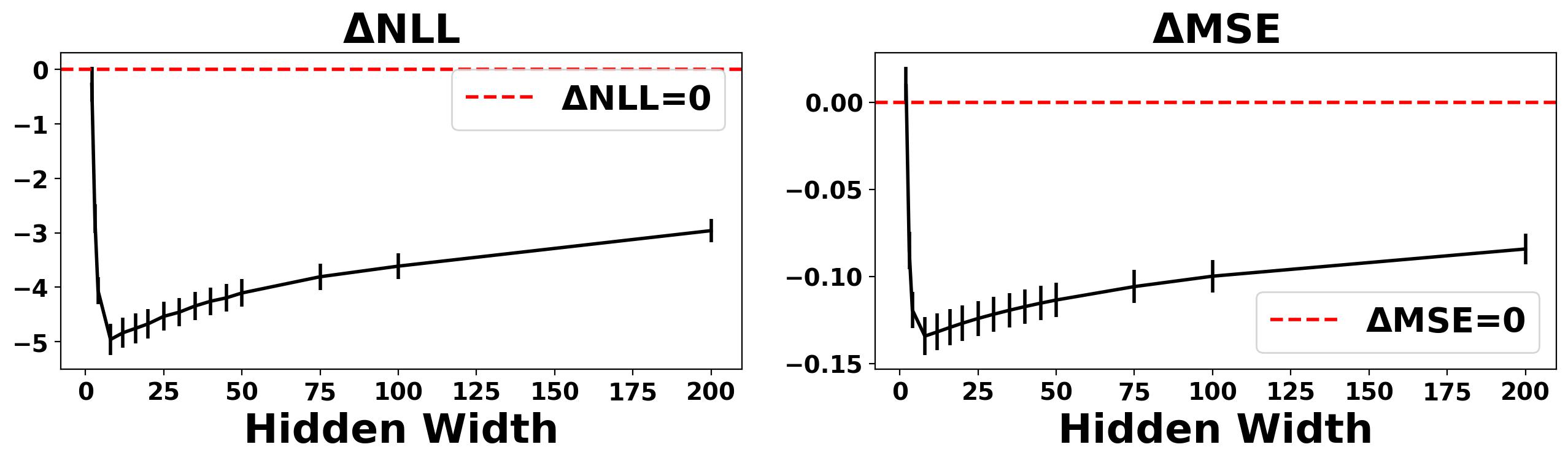}
         \caption{BNN with erf, $\sigma_\mathbf{W}^2=\sigma_\mathbf{b}^2=1.5$}
         \label{rbf_0.5_erf_1.5}
     \end{subfigure}
     \hfill
     \begin{subfigure}[b]{0.49\textwidth}
         \centering
         \includegraphics[width=\textwidth]{figs/data_fit/data_rbf_v_0.5_relu_v_2.0_trial_0.jpg}
         \caption{BNN with ReLU, $\sigma_\mathbf{W}^2=\sigma_\mathbf{b}^2=2.0$}
         \label{rbf_0.5_relu_2.0}
     \end{subfigure}
     \hfill
     \begin{subfigure}[b]{0.49\textwidth}
         \centering
         \includegraphics[width=\textwidth]{figs/data_fit/data_rbf_v_0.5_erf_v_2.0_trial_0.jpg}
         \caption{BNN with erf, $\sigma_\mathbf{W}^2=\sigma_\mathbf{b}^2=2.0$}
         \label{rbf_0.5_erf_2.0}
     \end{subfigure}
     \caption{Plots of the test performance of finite width BNNs:  The data is generated from GP-RBF with lengthscale $0.5$. We fit each dataset with BNNs with erf and ReLU nonlinearity. 
     $x$-axis denotes the hidden-layer width of the BNN; $y$-axis denotes 
     $\Delta \text{NLL}$, $\Delta \text{MSE}$ (Equation~\ref{eqn:delta_ll},\ref{eqn:delta_mse}).
     Being above/below the red dashed line means that finite width BNNs perform worse/better than the limiting NNGPs.
     We see that finite width BNNs can outperform the limiting NNGPs as the black curves are mostly below the red dashed line.}
     \label{fig:appdx_rbf_0.5}
\end{figure*}


\FloatBarrier
\subsection{Log Marginal Likelihood (LML)}\label{apdx:lml}
\FloatBarrier
 We provide additional results of the CDF of LML. Similar to Figure~\ref{marginal_log_ll_a},~\ref{marginal_log_ll_b}, we see that under different types of data generating models, BNNs with smaller hidden widths generate more diverse datasets than the limiting NNGP (the 
CDF of LML over datasets drawn from the data generating GP has more mass on the tail).
\begin{figure*}[htbp]
     \centering
       \begin{subfigure}[b]{0.49\textwidth}
         \centering
         \includegraphics[width=\textwidth]{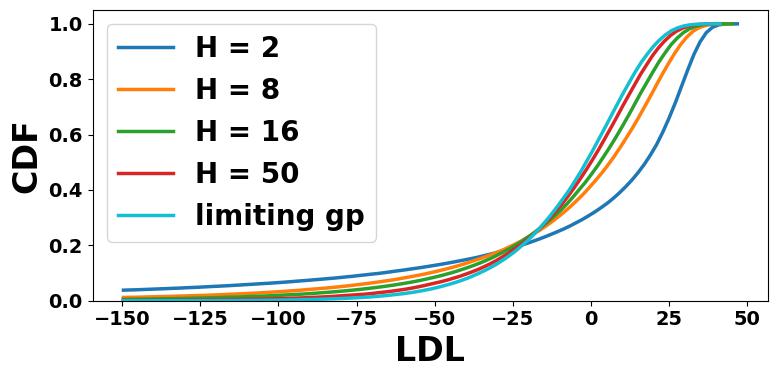}
         \caption{$\mathcal{D}\sim \text{GP-Arccos}(0.5)$; BNN with ReLU, $\sigma_\mathbf{W}^2=\sigma_\mathbf{b}^2=2.0$}
         \label{fig:appdx_lml_arccos_0.5}
     \end{subfigure}
     \hfill
    \begin{subfigure}[b]{0.49\textwidth}
         \centering
         \includegraphics[width=\textwidth]{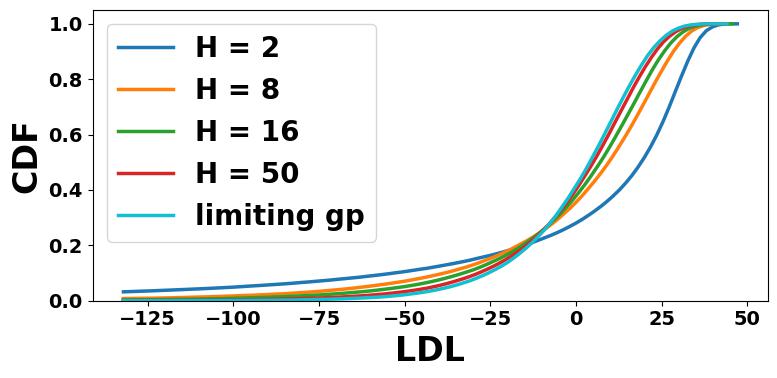}
         \caption{$\mathcal{D}\sim \text{GP-Arcsin}(0.5)$; BNN with erf, $\sigma_\mathbf{W}^2=\sigma_\mathbf{b}^2=2.0$}
    \label{fig:appdx_lml_arcsin_0.5}
     \end{subfigure}
     \caption{CDF of LML (Eqn~\ref{eqn:lml_gp}) with datasets $\mathcal{D}$ sampled from BNNs with different hidden widths: ~\ref{fig:appdx_lml_arccos_0.5} $\mathcal{D}\sim$GP-Arccos(0.5);~\ref{fig:appdx_lml_arcsin_0.5} $\mathcal{D}\sim$GP-Arcsin(0.5). $x$-axis denotes the values of LML while $y$-axis denotes CDF. Each color represents either a BNN with a specific hidden width or the limiting NNGP. BNNs with smaller hidden widths generate more datasets with higher LML and lower LML than the limiting GP at the same time.}
\end{figure*} 

\subsection{Low-pass filtered BNNs} \label{apdx:lpf-bnn-vs-bnn}
\begin{figure*}[t]
     \centering
     \begin{subfigure}[b]{0.49\textwidth}
         \centering
         \includegraphics[width=\textwidth]{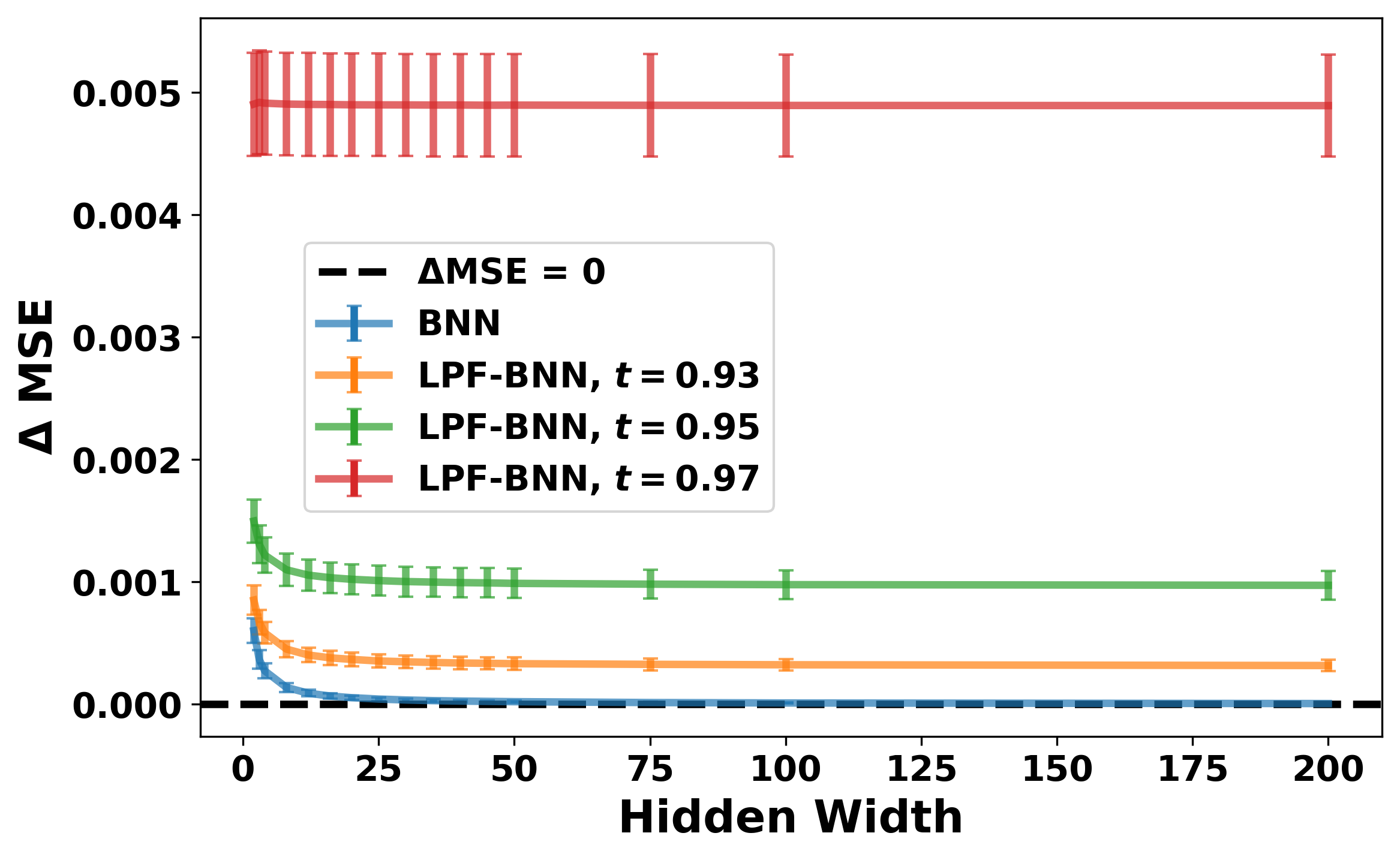}
         \caption{BNN with ReLU, $\sigma_\mathbf{W}^2=\sigma_\mathbf{b}^2=0.5$}
     \end{subfigure}
     \hfill
    \begin{subfigure}[b]{0.49\textwidth}
         \centering
         \includegraphics[width=\textwidth]{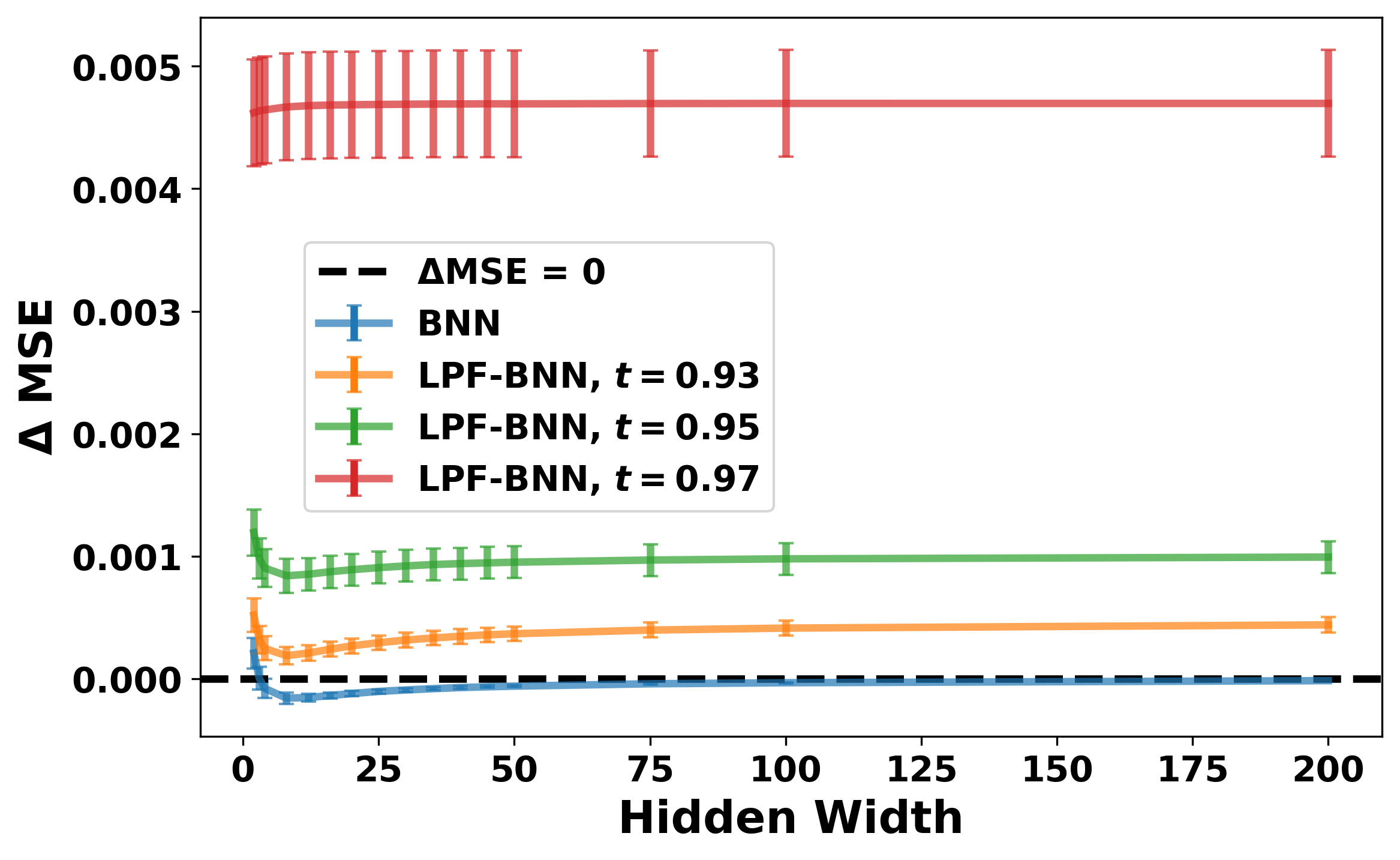}
         \caption{BNN with ReLU, $\sigma_\mathbf{W}^2=\sigma_\mathbf{b}^2=1.0$}
     \end{subfigure}
     
     \begin{subfigure}[b]{0.49\textwidth}
         \centering
         \includegraphics[width=\textwidth]{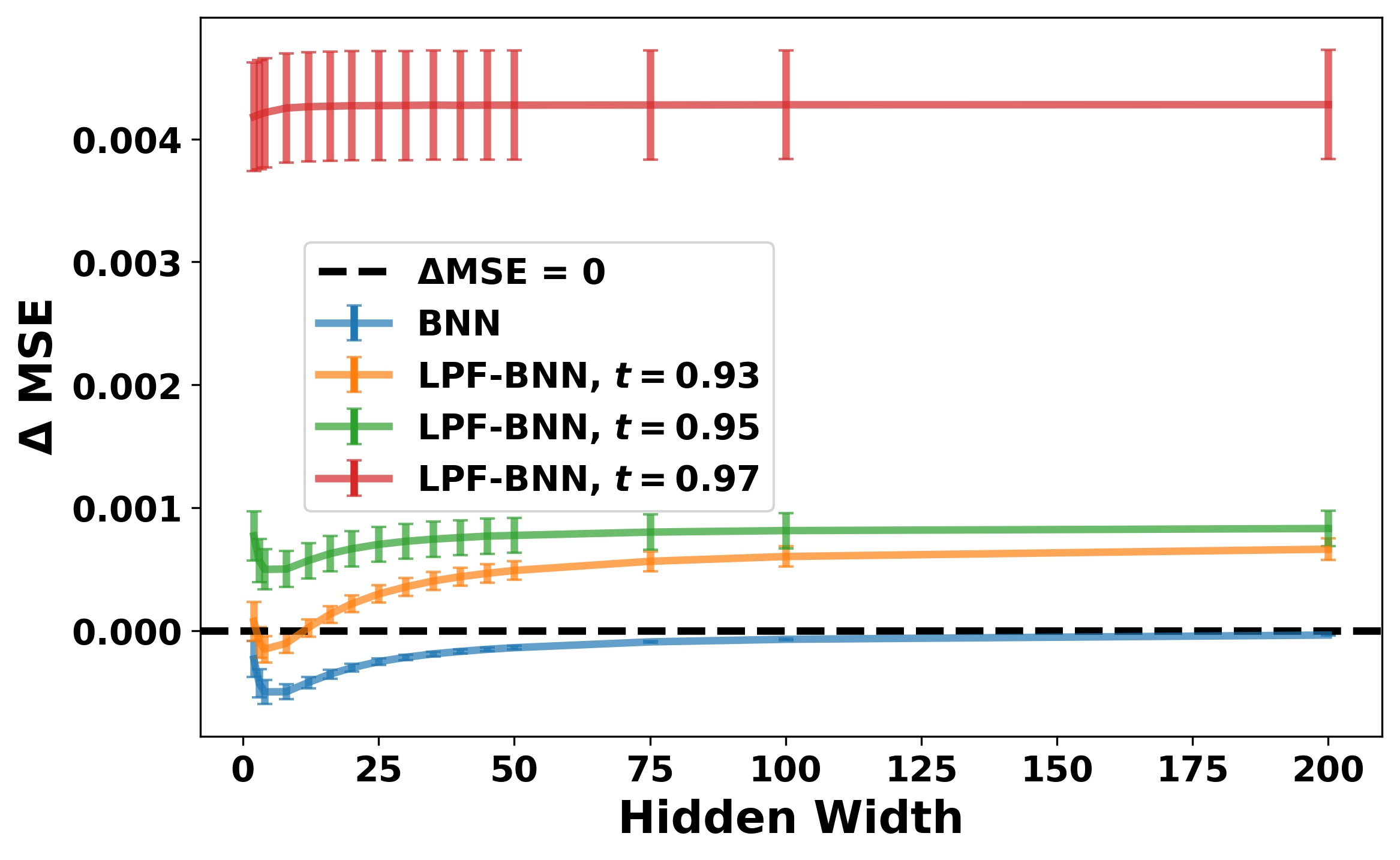}
         \caption{BNN with ReLU, $\sigma_\mathbf{W}^2=\sigma_\mathbf{b}^2=1.5$}
     \end{subfigure}
     \hfill
     \begin{subfigure}[b]{0.49\textwidth}
         \centering
         \includegraphics[width=\textwidth]{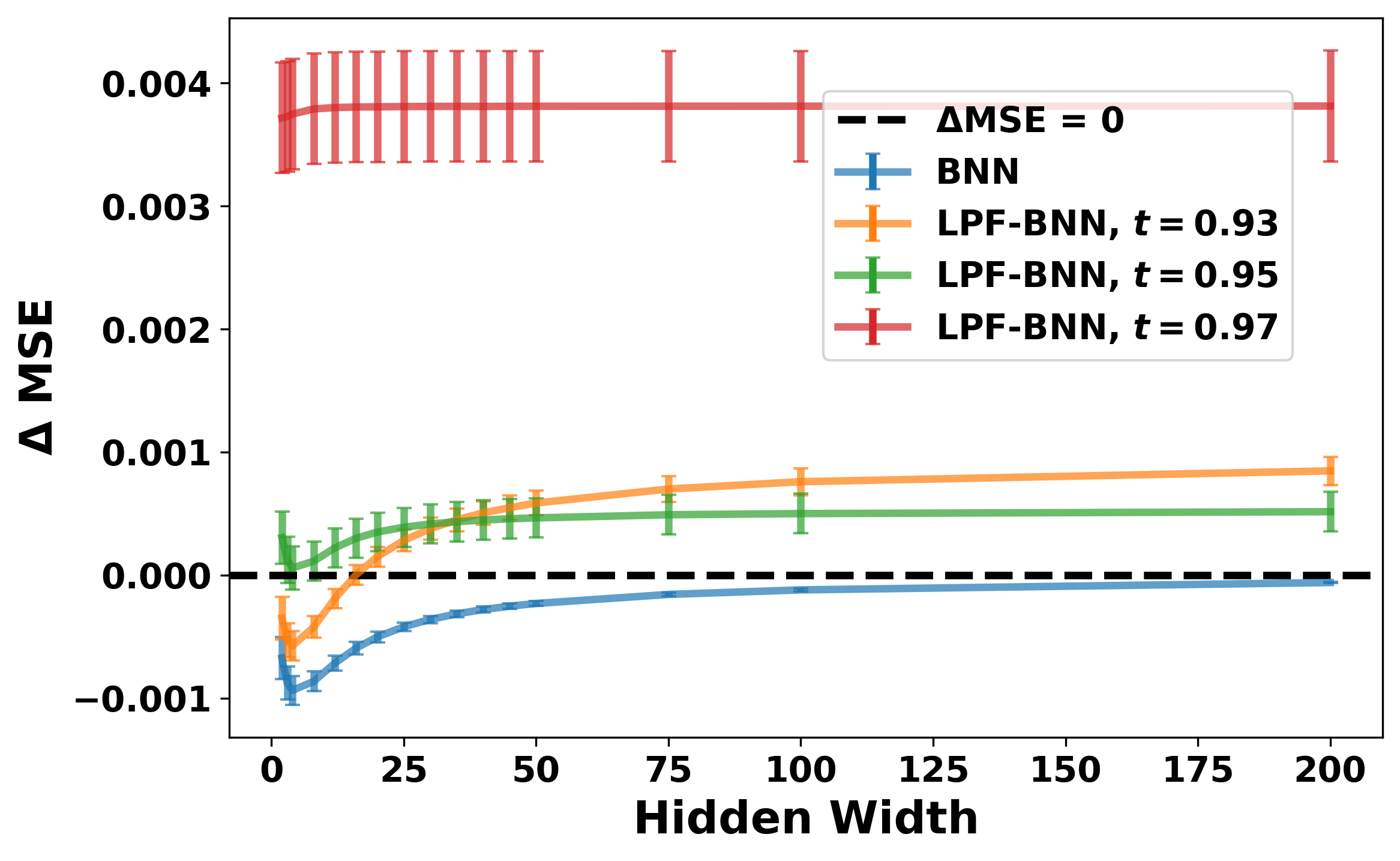}
         \caption{BNN with ReLU, $\sigma_\mathbf{W}^2=\sigma_\mathbf{b}^2=2.0$}
     \end{subfigure}
     \caption{$\Delta_\text{MSE}$ of BNNs and lowepass-filtered BNNs, as a function of hidden width $H$ ($\Delta \text{MSE} < 0$ means a worse performance than the limiting NNGP). Higher thresholds $t$ correspond to more high frequencies removed. The performance is averaged over 200 datasets drawn from a $\text{GP-Arccos}(0.5)$. Error bars represent standard errors. Thresholds were determined experimentally ($t \leq 0.91$ was found not to alter the posterior predictive).}
     \label{fig:lpf_bnn_arccos_arccos}
\end{figure*}

\begin{figure*}[tb]
     \centering
     \begin{subfigure}[b]{0.49\textwidth}
         \centering
         \includegraphics[width=\textwidth]{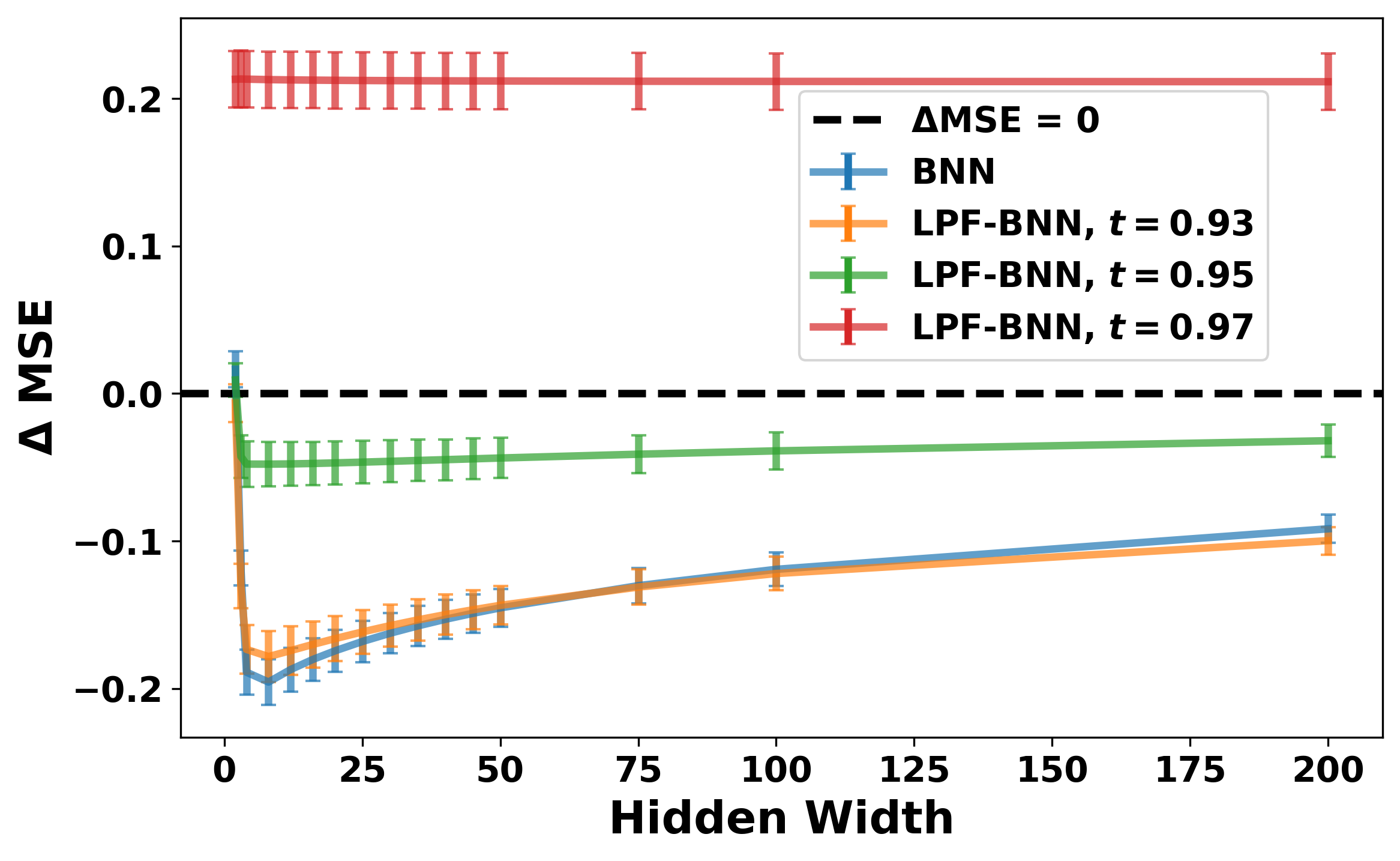}
          \caption{BNN with ReLU, $\sigma_\mathbf{W}^2=\sigma_\mathbf{b}^2=0.5$}
     \end{subfigure}
     \hfill
    \begin{subfigure}[b]{0.49\textwidth}
         \centering
         \includegraphics[width=\textwidth]{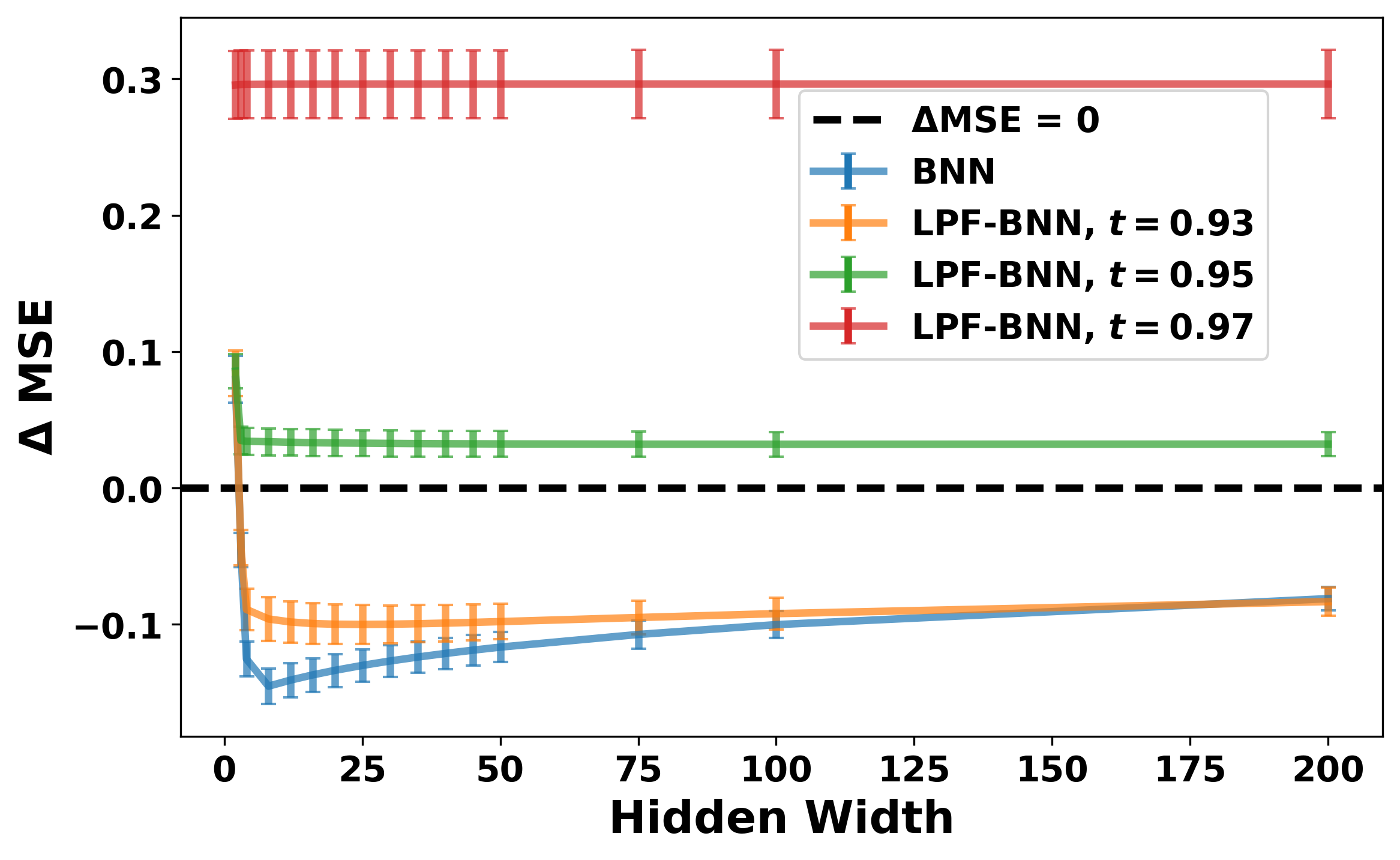}
       \caption{BNN with ReLU, $\sigma_\mathbf{W}^2=\sigma_\mathbf{b}^2=1.0$}
     \end{subfigure}
     
     \begin{subfigure}[b]{0.49\textwidth}
         \centering
         \includegraphics[width=\textwidth]{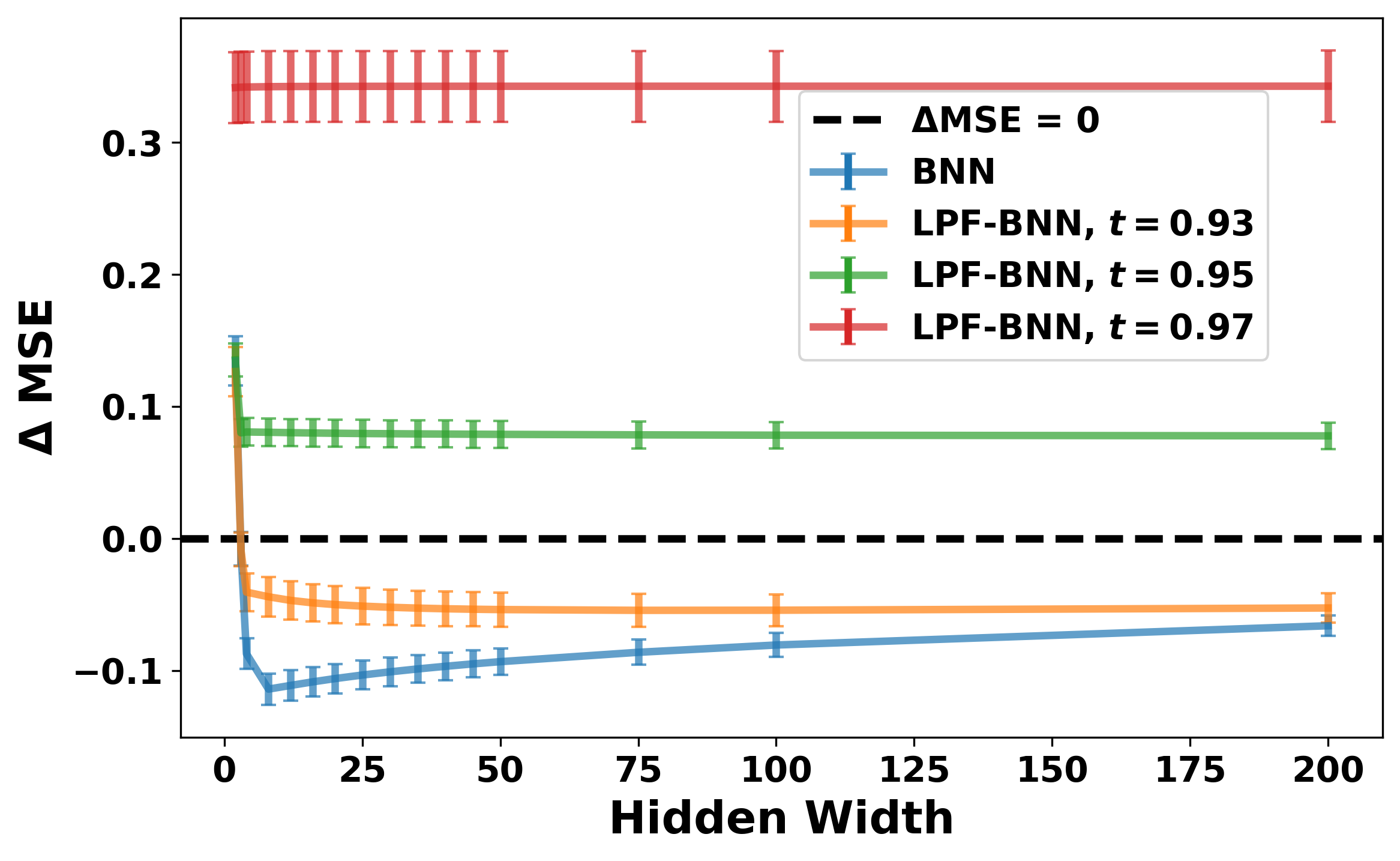}
          \caption{BNN with ReLU, $\sigma_\mathbf{W}^2=\sigma_\mathbf{b}^2=1.5$}
     \end{subfigure}
     \hfill
     \begin{subfigure}[b]{0.49\textwidth}
         \centering
         \includegraphics[width=\textwidth]{figs/lpf_bnn/rbf_v_0.5_arccos_v_2.0_mse_test.png}
          \caption{BNN with ReLU, $\sigma_\mathbf{W}^2=\sigma_\mathbf{b}^2=2.0$}
     \end{subfigure}
     \caption{Analogous to Figure \ref{fig:lpf_bnn_arccos_arccos} but with data generated from a GP with an RBF kernel, $\mathcal{D}\sim$ GP-RBF(0.5) (so the limiting NNGP has an Arccos kernel).}
     \label{fig:lpf_bnn_rbf_arccos}
\end{figure*}

\begin{figure*}[tb]
     \centering
     \begin{subfigure}[b]{0.49\textwidth}
         \centering
         \includegraphics[width=\textwidth]{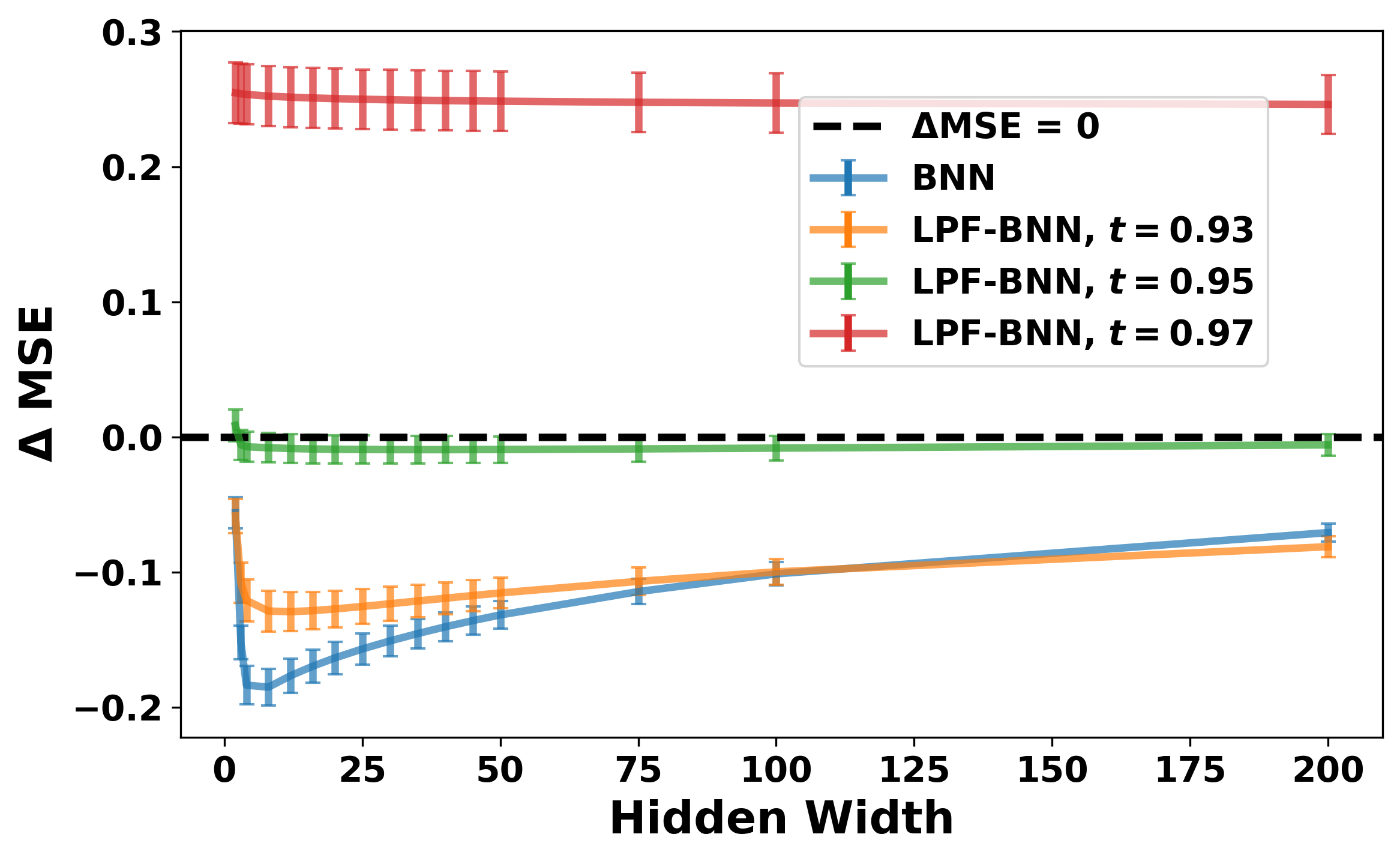}
    \caption{BNN with erf, $\sigma_\mathbf{W}^2=\sigma_\mathbf{b}^2=0.5$}     \end{subfigure}
     \hfill
    \begin{subfigure}[b]{0.49\textwidth}
         \centering
         \includegraphics[width=\textwidth]{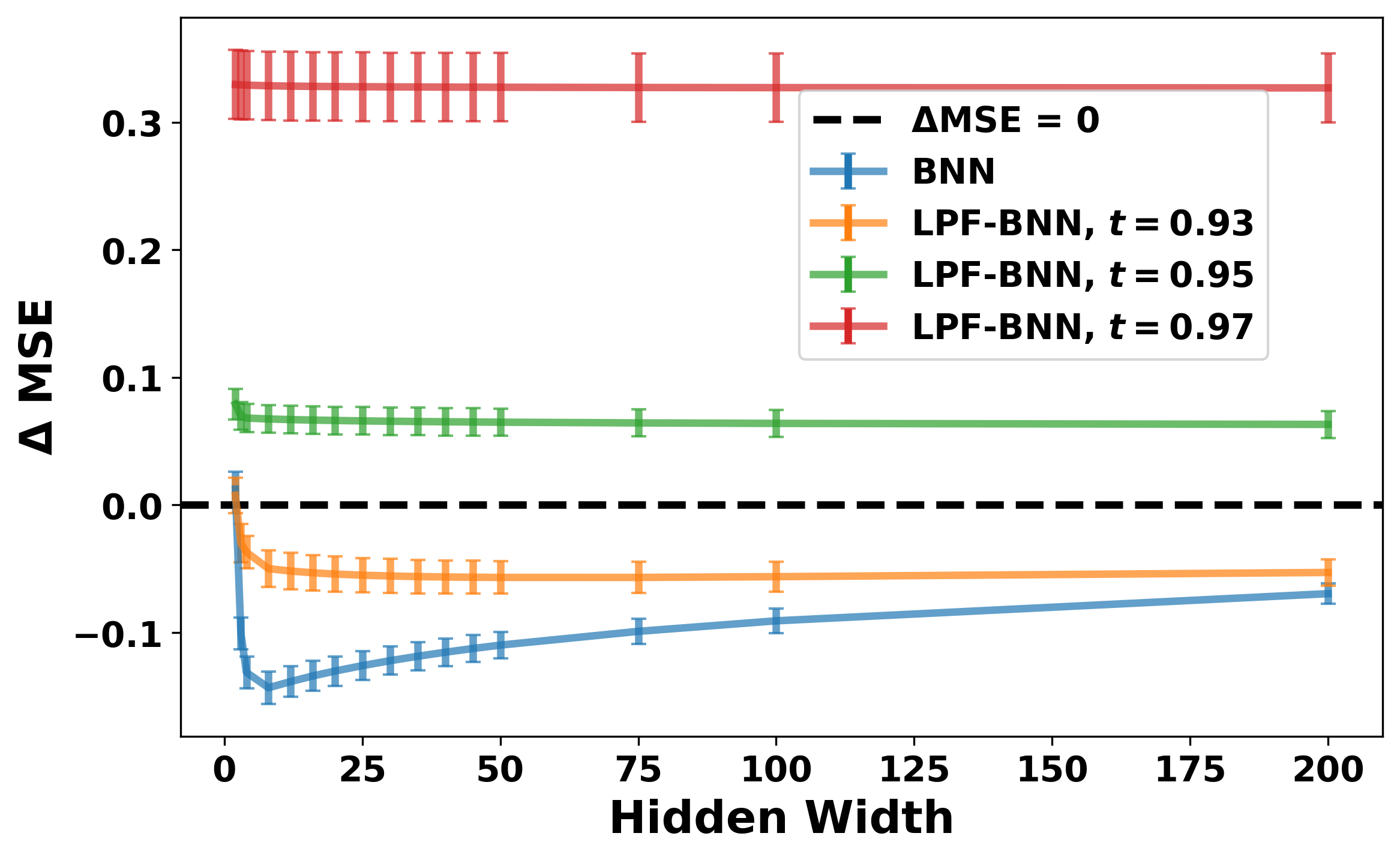}
    \caption{BNN with erf, $\sigma_\mathbf{W}^2=\sigma_\mathbf{b}^2=1.0$}        \end{subfigure}
     
     \begin{subfigure}[b]{0.49\textwidth}
         \centering
         \includegraphics[width=\textwidth]{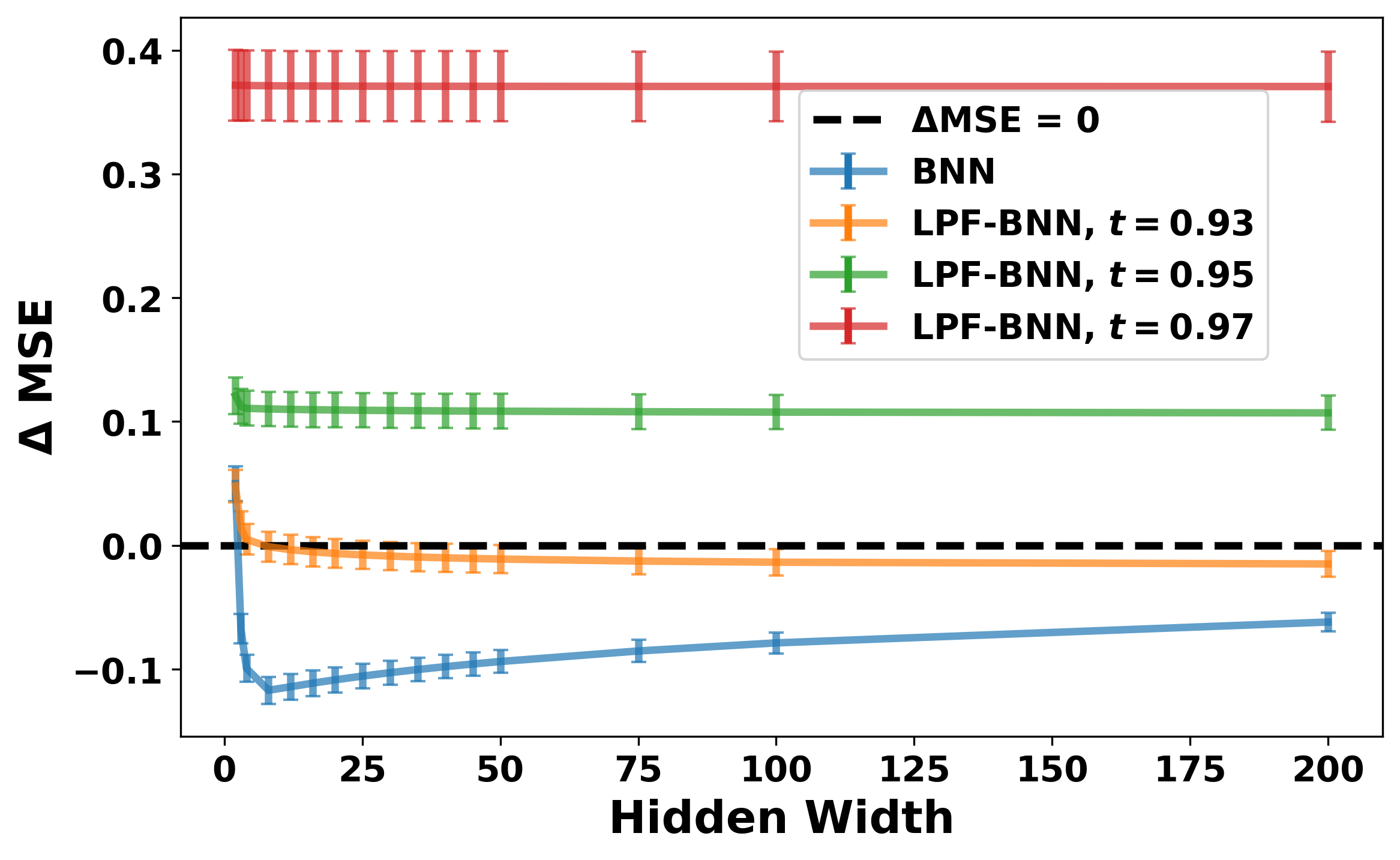}
    \caption{BNN with erf, $\sigma_\mathbf{W}^2=\sigma_\mathbf{b}^2=1.5$}        \end{subfigure}
     \hfill
     \begin{subfigure}[b]{0.49\textwidth}
         \centering
         \includegraphics[width=\textwidth]{figs/lpf_bnn/rbf_v_0.5_arcsin_v_2.0_mse_test.png}
    \caption{BNN with erf, $\sigma_\mathbf{W}^2=\sigma_\mathbf{b}^2=2.0$}        \end{subfigure}
     \caption{Analogous to Figure \ref{fig:lpf_bnn_arccos_arccos} but with data generated from a GP with an RBF kernel, $\mathcal{D}\sim$ GP-RBF(0.5), and a BNN with erf activation (so the limiting NNGP has an Arcsin kernel).}
     \label{fig:lpf_bnn_rbf_arcsin}
\end{figure*}

\paragraph{Removing high-frequency components from the BNN hurts its ability to adapt to model mismatch.}
We fix the data-generating process to be that of a GP.
Across varying widths, we then compare the $\Delta_\text{MSE}$ (Equation~\ref{eqn:delta_mse}) of a BNN with that of low-pass filtered BNNs with different amounts of filtering, $t \in [0, 1]$.
As expected, as the width increases, $\Delta_\text{MSE}$ of the BNN converges to $0$.
Moreover, under no model mismatch (i.e., when the limiting NNGP model matches the data generating process, as in Figure \ref{fig:lpf_bnn_arccos_arccos}a), the BNN always under-performs the limiting NNGP.
In contrast, under model mismatch (remainder of Figure \ref{fig:lpf_bnn_arccos_arccos}, as well as Figures \ref{fig:lpf_bnn_rbf_arccos} and \ref{fig:lpf_bnn_rbf_arcsin}), small-width BNNs always outperform the NNGP (i.e., $\Delta \text{MSE} < 0$). 
As we hypothesized, removing the high-frequency components from prior functions hinders the BNN's ability to adapt to model mismatch. 

\begin{figure*}[p]
    \centering     
    \includegraphics[width=\textwidth]{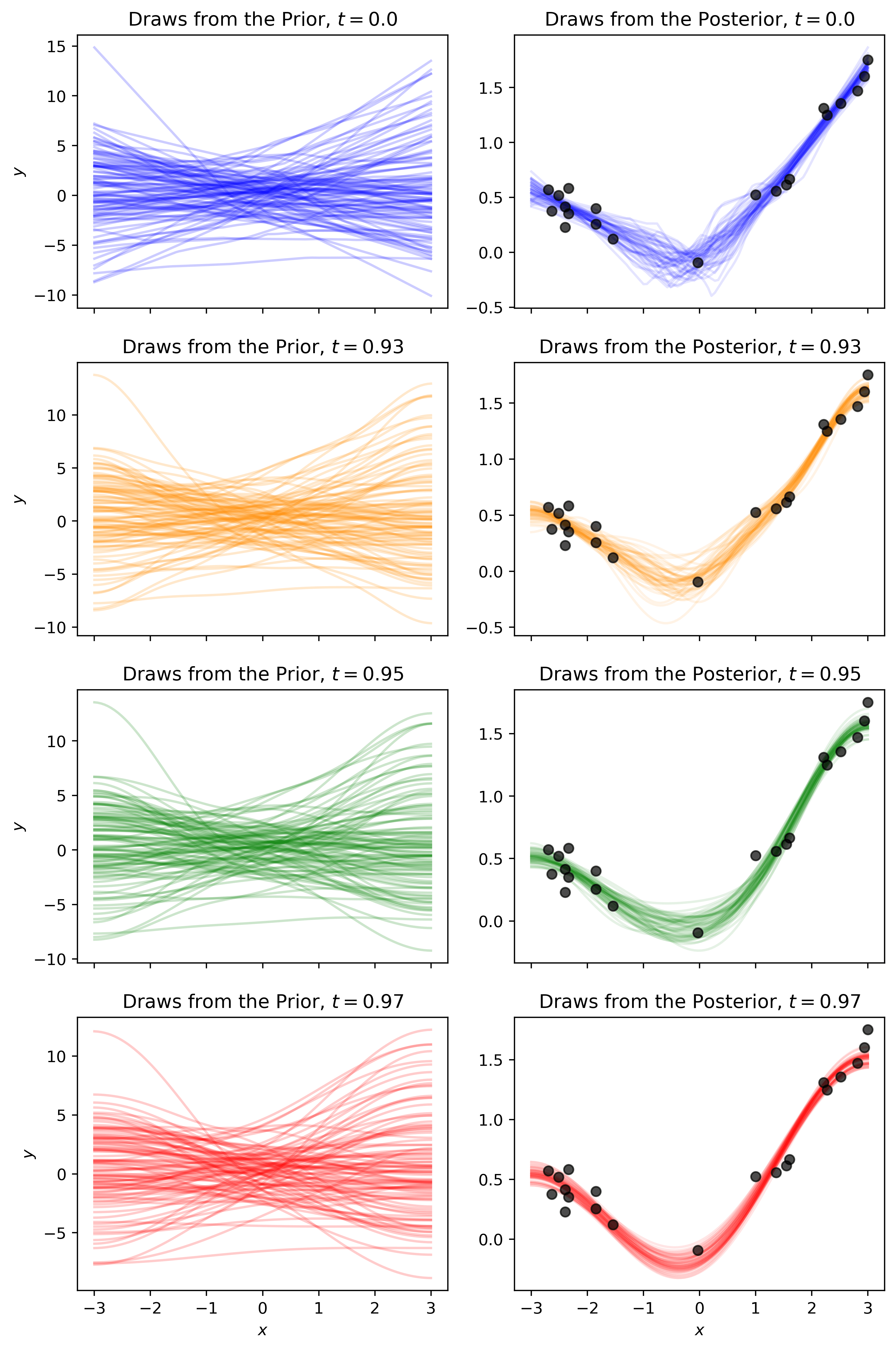}
    \caption{A BNN with a width of $15$, a prior weight variance of $2.0$ and a ReLU activation, trained on data drawn from a $\text{GP-Arccos}(0.5)$. As the filtering threshold $t$ increases, the functions under both the prior and posterior become less wiggly.}
    \label{fig:arccos-arccos-qualitative}
\end{figure*}

\begin{figure*}[p]
    \centering     
    \includegraphics[width=\textwidth]{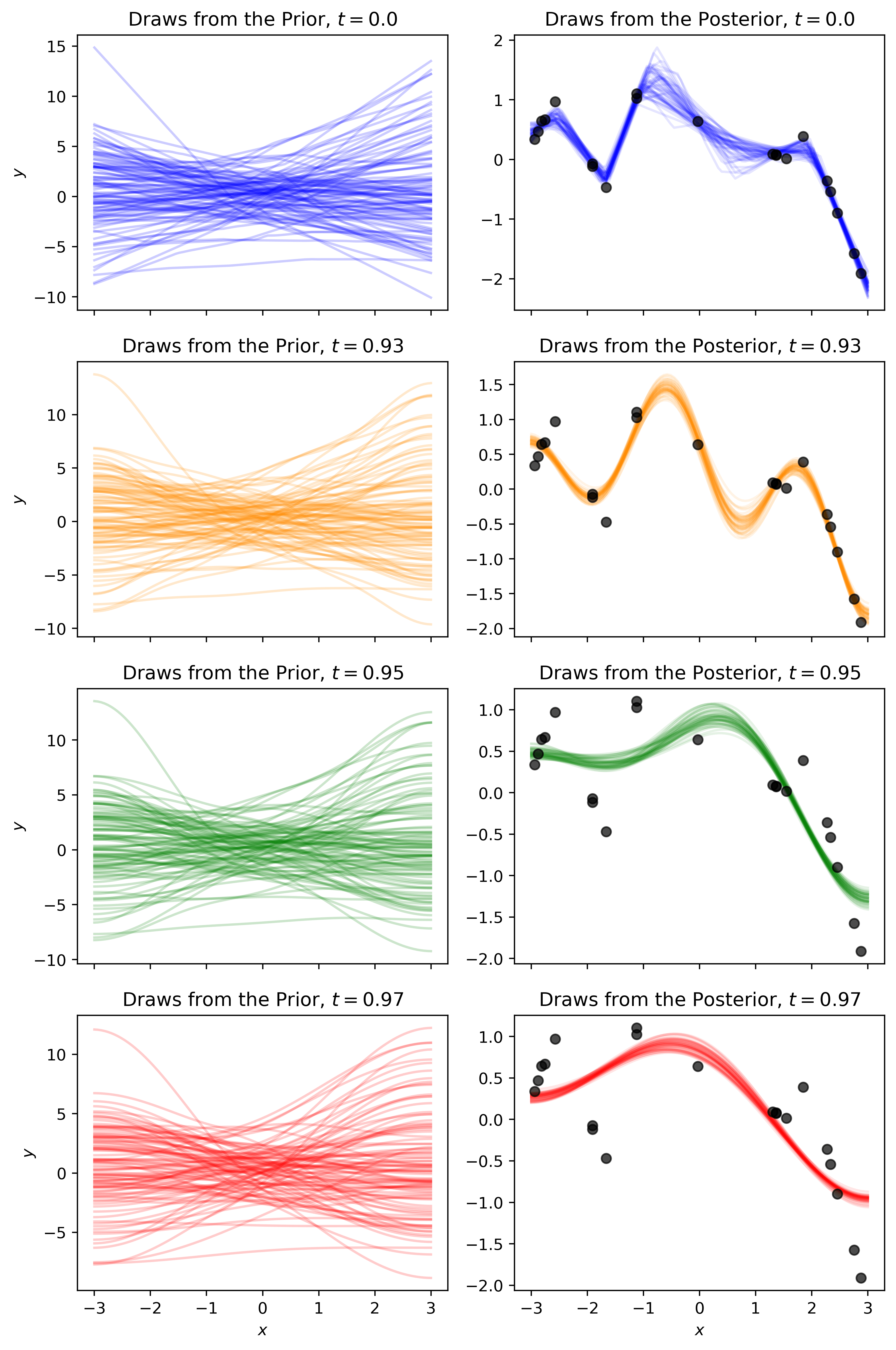}
    \caption{A BNN with a width of $15$, a prior weight variance of $2.0$ and a ReLU activation, trained on data drawn from a $\text{GP-RBF}(0.5)$. As the filtering threshold $t$ increases, the functions under both the prior and posterior become less wiggly.}
    \label{fig:rbf-arccos-qualitative}
\end{figure*}

\paragraph{A qualitative comparison of a BNN vs. a Low-pass filtered BNN.} Figures \ref{fig:arccos-arccos-qualitative} and \ref{fig:rbf-arccos-qualitative} show how a BNN ($t = 0$) differs from a low-pass filtered BNN ($t > 0$), both in its prior and in its posterior predictive distributions. Specifically, as $t$ increases, the functions being drawn become more smooth.

\end{document}